\pdfoutput=1
\documentclass[11pt]{article}

\usepackage[preprint]{acl}

\usepackage{times}
\usepackage{latexsym}

\usepackage[T1]{fontenc}
\usepackage[utf8]{inputenc}

\usepackage{microtype}

\usepackage{inconsolata}
\usepackage{graphicx}
\usepackage{color}
\usepackage{multirow}
\usepackage{amsmath}
\usepackage{array}
\usepackage{booktabs}
\usepackage{float} 
\usepackage{arydshln}
\usepackage{subcaption}
\usepackage{xspace}
\usepackage{makecell}
\usepackage{url}
\usepackage{pifont}
\usepackage[colorinlistoftodos]{todonotes}

\newcommand{\model}{\textsc{SG-KBQA}\xspace}

\newcommand{\info}[1]{}

\title{Beyond Seen Data: Improving KBQA Generalization Through Schema-Guided Logical Form Generation}

\author{
  Shengxiang Gao  \hspace{10mm} Jey Han Lau  \hspace{10mm} Jianzhong Qi\thanks{Corresponding author.} \vspace{3mm} \\
  The University of Melbourne \\
  \texttt{shengxiang.gao1@student.unimelb.edu.au} \\ 
  \texttt{\{laujh, jianzhong.qi\}@unimelb.edu.au}\\
}

\begin{document}
\maketitle

% \footnotetext[1]{Corresponding author.}

\begin{abstract}
Knowledge Base Question Answering (KBQA) aims to answer user questions in natural language using rich human knowledge stored in large KBs. As current KBQA methods struggle with unseen knowledge base elements and their novel compositions at test time, we introduce \textbf{\model} --- a novel model that injects schema contexts into entity retrieval and logical form generation to tackle this issue. 
It exploits information about the semantics and structure of the knowledge base provided by schema contexts to enhance generalizability. We show that \model\ achieves strong generalizability, outperforming state-of-the-art models on three commonly used benchmark datasets across a variety of test settings. 
Our source code is available at \url{https://github.com/gaosx2000/SG_KBQA}.
%\url{https://anonymous.4open.science/r/SG-KBQA-7895}. 
%Our source code is available at 
\end{abstract}

\section{Introduction}\label{sec:introduction}

Knowledge Base Question Answering (KBQA) aims to answer user questions expressed in natural language with information from a {knowledge base}~(KB). This offers user-friendly access to rich human knowledge from large KBs such as Freebase~\citep{bollacker_freebase_2008}, DBPedia~\citep{auer_dbpedia_2007} and Wikidata~\citep{vrandecic_wikidata_2014}, and it has broad applications in QA~\citep{zhou_commonsense_2018}, recommendation~\citep{guo_survey_2022}, and information retrieval~\citep{jalota_lauren_2021}.

Semantic Parsing (SP) has been shown to be an effective method for KBQA, where the core idea is to translate the input natural language question into a structured logical form (e.g., SPARQL or S-Expression~\citep{gu_beyond_2021}), which is then executed to yield the question answer. 

\begin{figure}[t]
    \centering
    \includegraphics[width=\columnwidth]{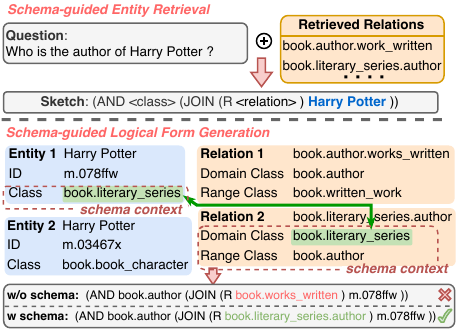}
    \caption{Schema-guided entity retrieval (top) and logical form generation (bottom). Green arrows and boxes highlight schema-level connections (overlapping classes) among KB elements. ``w schema'' and ``w/o schema'' logical forms  denote whether the composition of KB elements adheres to the KB schema, respectively.}
    \label{fig:intro_example}
\end{figure}

A key challenge here is to learn a mapping between mentions of entities and relations in the input question to corresponding KB elements to form the logical form. Given a large number of entities and relations in ambiguous surface forms, and the flexibility in questions expressed in natural language, this mapping process typically yields a set of candidate entities (relations) for each mentions of entities (relations). \emph{The challenge then becomes to uncover the right composition of entities and relations from the sets of candidates.}

Figure~\ref{fig:intro_example} shows an example. Two entities named \textsf{Harry Potter} (a book series and a main  character in them) and two authorship relations of similar names were identified as candidates. Combining the top-ranked entity (the book series) with the top-ranked relation \textsf{book.author.works\_written} (book authorship) yields an invalid and unexecutable logical form, as the entity is a book series, not a book. 

Due to the vast number of KB elements and their compositions, it is difficult (if not impossible) to train a model with all feasible compositions of KB elements that might be queried. For example, Freebase~\citep{bollacker_freebase_2008} has over 39 million entities, 8,000 relations, and 4,000 classes. Furthermore, some KBs (e.g., NED~\citep{mitchell_ned_2018}) are not static as they continue to grow. When a KB element composition is unseen at training, errors like the example above may occur. 

A few studies consider model generalizability to non-I.I.D. settings, where the test set contains schema items (i.e., relations, classes and functions) or their compositions that are unseen during training (i.e., \emph{zero-shot} and \emph{compositional generalization}, respectively). They propose  retrieval methods to retrieve KB elements or compositions more relevant to input questions, and use them to construct logical forms~\citep{shu_tiara_2022, gu_dont_2023}. 
    
Despite these efforts, achieving compositional and zero-shot generalization remains challenging: (1)~\textbf{Entity retrieval is a bottleneck.} Current entity retrieval methods often fail to accurately detect entities mentioned in questions containing schema items unseen at training. This is because such items introduce novel contextual patterns in questions, making it difficult to identify the correct boundary of entity mentions. 
The resulting entity retrieval errors propagate and lead to errors in the logical forms generated. (2)~\textbf{Schema-level connection between KB elements have been missed.} Existing methods are not explicitly trained to capture compositions of KB elements that are feasible based on their schema. Instead, they tend to reproduce KB element compositions observed at training, making them difficult to generalize to unseen compositions.

To address these challenges, we propose a \underline{s}chema-\underline{g}uided model for KBQA (SG-KBQA), that incorporates KB schema to guide both entity retrieval and logical form generation. Unlike previous approaches that initiate the pipeline with entity retrieval, \model adopts a schema-first principle, prioritizing schema understanding as the foundation for downstream logical form generation. 

\model begins with relation retrieval, employing a pre-trained language model (PLM)-based~\citep{devlin_bert_2019} retriever to retrieve top-ranked relations from the KB that are most relevant to the input question. Benefiting from the generalization capability of pre-trained language models (and that there are much fewer relations and their surface form variants than entities in a KB), relations that are semantically similar to the question---yet unseen at training---can still be included among the top-ranked retrieved relations (as validated in our study). Then, as illustrated in Figure~\ref{fig:intro_example}, we introduce a schema-guided entity retrieval (SER) module. This module employs a logical form sketch parser that converts the input question and retrieved relations into logical form sketches by a Seq2Seq model. These top-ranked relations provide schema context that is relevant to the question but not observed at training, hence helping the model distinguish actual entity mentions from unseen schema items in the question. More precise entity mentions are then extracted from the generated sketches, thereby improving the zero-shot generalizability of entity retrieval.

To further mitigate error propagation between the retrieval and generation stages, we defer entity disambiguation to the logical form generation stage. For each entity mention, all top-ranked matched candidates are retained as candidate entities for the preceding generation stage. 

Further exploiting the schema-guided idea, we propose a schema-guided logical form generation (SLFG) module that fine-tunes a large language model (LLM) to reason over feasible compositions of KB elements based on their underlying schema contexts. As Figure~\ref{fig:intro_example} shows, we feed the input question, the retrieved candidate relations and entities, plus their corresponding schema contexts, i.e., the (domain and range) classes of the relations and entities, into the LLM for logical form generation. The domain and range classes of a relation refer to the classes to which its subject and object entities belong. \emph{Together with the class of the candidate entities, they provide explicit training signals to guide the LLM to look for KB elements that can be connected together (and hence are more likely to form executable logical forms).}
As a result, our SLFG module generalizes to compositions of KB elements unseen at training.

To summarise: (1)~We introduce \model\ to solve the KBQA problem under non-I.I.D. settings, where test input contains unseen schema items or their compositions during training.
 (2)~We introduce schema-guided modules for entity retrieval and logical form generation with deferring entity disambiguation to enhance both compositional and zero-shot generalization. These modules can also be incorporated into existing SP-based KBQA systems to improve their generalization performance.  %in non-I.I.D. scenarios. 
(3)~We conduct experiments on three popular benchmark datasets and find \model\ outperforming SOTA models.
In particular, on non-I.I.D GrailQA our model tops all three leaderboards for the overall, zero-shot, and compositional generalization settings, outperforming SOTA models by 3.3\%, 2.9\%, and 4.0\% (F1) respectively.

\section{Related Work}\label{sec:literature}

\paragraph{Knowledge Graph Question Answering}
Early KBQA solutions can be widely categorized as {Information Retrieval-based} (IR-based)~\citep{he_improving_2021, zhang_subgraph_2022} or {Semantic Parsing-based} (SP-based) solutions~\cite{cao_program_2022, ye_rng-kbqa_2022}.

%IR-based methods construct a question-specific subgraph starting from the topic entities. They then reason over the subgraph to derive the answer. SP-based methods focus on transforming input questions into logical forms, which re then executed to retrieve answers~\citep{shu_tiara_2022, ye_rng-kbqa_2022}.  
% Benefiting from the strong natural language understanding and logical form generation capabilities of LLMs, recent LLM-based KBQA methods that follow the SP-based paradigm achieve promising results~\cite{luo_chatkbqa_2024, luo_ReasoningOnGraph_2024}. Meanwhile, a branch of works leverages LLMs through in-context learning or fine-tune to perform step-by-step reasoning or tool invocation over a question-specific subgraph, typically rooted at the given topic entity and constrained within a certain hops~\citep{sun_ThinkonGraph_2024, ma_ThinkonGraph2_2025, jiang_KGAgent_2024}. 

Benefiting from the strong natural language understanding and reasoning abilities of LLMs, recent LLM-based KBQA methods have achieved promising results. A branch of work employs LLMs to produce reasoning trajectories prior to producing final answers. For example, \citet{luo_ReasoningOnGraph_2024} and \citet{luo_gcr_2025} fine-tune LLMs to directly generate the reasoning trajectories in a single pass. In contrast, \citet{sun_ThinkonGraph_2024}, \citet{liu_icsu_2024} and \citet{sui_FiDeLiSFaithfulReasoning_2025} follow an agentic framework, where the model performs step-by-step tool invocations to iteratively traverse the subgraph starting from topic entities, assuming that topic entities are given.

% \sxfix{A branch of work employs LLMs to produce reasoning trajectories or follow an agentic framework that performs step-by-step tool invocations over question-specific subgraphs to obtain question answers, assuming that topic entities are given~\citep{sun_ThinkonGraph_2024, luo_ReasoningOnGraph_2024,sui_FiDeLiSFaithfulReasoning_2025, liu_icsu_2024}.}

% A branch of work employs LLMs to produce reasoning trajectories or use LLMs as an agent to perform step-by-step tool invocation over a question-specific subgraph to obtain question answers, assuming that topic entities are given~\citep{sun_ThinkonGraph_2024, luo_ReasoningOnGraph_2024,sui_FiDeLiSFaithfulReasoning_2025, liu_icsu_2024}.

Others follow the SP-based paradigm, using LLMs to generate approximate logical form sketches through few-shot in-context learning or fine-tuning~\citep{cao_program_2022,li_few-shot_2023,li_flexkbqa_2024,luo_chatkbqa_2024, wang_no_2024}. The inaccurate or ambiguous KB elements in the generated sketches are further refined through a retrieval stage, aligning them with actual KB elements to construct complete logical forms.

However, these methods often fail to generalize over test questions that refer to KB elements or their compositions unseen during training, or when the topic entities are not known. Our \model improves KBQA generalizability through a schema-guided approach. While we also use LLMs to generate logical form sketches, we incorporate retrieved relations to guide sketch generation for entity mention extraction, thereby improving the generalizability of entity retrieval, while \emph{we do not refine these sketches to produce the final output logical forms}.

\paragraph{KBQA under Non-I.I.D. Settings}

Studies considering non-I.I.D. settings can be largely classified into \emph{ranking-based} and \emph{generation-based} methods. 

Ranking-based methods~\cite{gu_beyond_2021,gu_dont_2023} start from retrieved entities, traverse the KB, and construct the target logical form by ranking the traversed paths. %\citet{gu_beyond_2021} enumerate and rank all possible logical forms within two hops  of retrieved entities, while \citet{gu_dont_2023} incrementally expand and rank paths from retrieved entities. 

Generation-based methods transform an input question into a logical form using a Seq2Seq model (e.g., T5~\cite{raffel_exploring_2023}).
They often use additional contexts beyond the question to augment the input of the Seq2Seq model and enhance its generalizability. For example,~\citet{ye_rng-kbqa_2022} use  the top-5 candidate logical forms enumerated from the retrieved entities. 
\citet{shu_tiara_2022} further use top-ranked relations, \emph{disambiguated entities}, and classes (retrieved \emph{separately}). \citet{zhang_fc-kbqa_2023} use connected pairs of retrieved KB elements. 

Our \model\ adopts a generation-based approach, training the LLM to reason over candidate entities and relations, using their schema contexts (i.e., classes) to infer connectivity. 
This enables the model to compose novel logical forms without seeing them at training, hence generalizing better in larger, noisier search spaces. Additionally, we defer entity disambiguation to the generation stage, mitigating error propagation caused by early disambiguation without context.

\paragraph{KBQA Entity Retrieval}
KBQA entity retrieval typically has three steps: {entity mention detection}, {candidate entity retrieval}, and {entity disambiguation}. BERT~\cite{devlin_bert_2019}-based named entity recognition  is used for entity mention detection from input questions. To retrieve KB entities for the entity mentions, the FACC1 dataset~\cite{gabrilovich_facc1_2013} is often used, with over 10 billion surface forms and popularity scores of Freebase entities. \citet{gu_beyond_2021} use  popularity scores for entity disambiguation, while \citet{ye_rng-kbqa_2022} and \citet{shu_tiara_2022} adopt a BERT reranker.

\section{Preliminaries}\label{sec:preliminary}

\begin{figure*}[ht]
    \centering
    \includegraphics[width=1\linewidth]{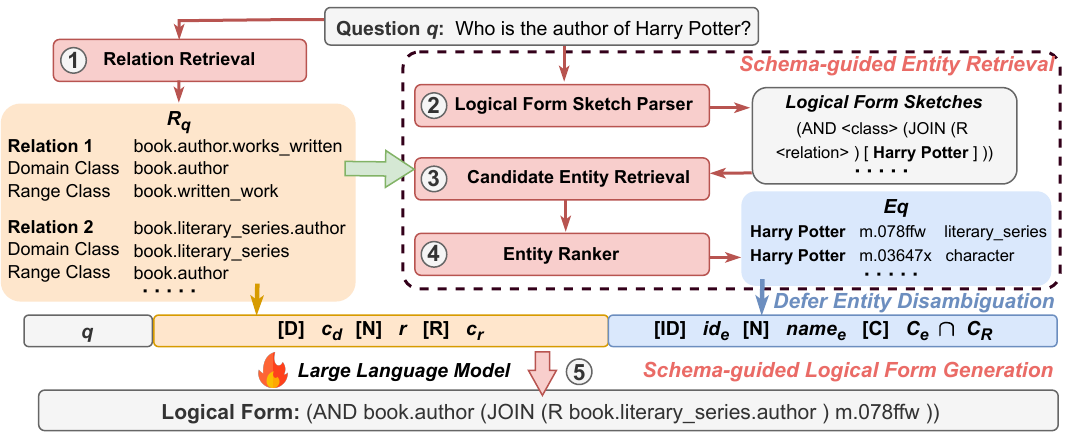}
    \caption{Overview of \model. The model consists of two novel modules: \emph{schema-guided entity retrieval} (SER) and \emph{schema-guided logical form generation} (SLFG). 
    Given a question $q$, the model first retrieves and ranks candidate relations (\textcircled{1}). 
    In SER, $q$ and the top-ranked relations $R_q$ are used to generate logical form sketches and extract entity mentions (\textcircled{2}). 
    Based on these mentions and $R_q$, the model retrieves and ranks candidate entities~(\textcircled{3}), producing the top entities $E_q$ (\textcircled{4}). 
    Entity disambiguation is deferred by directly passing $E_q$ to SLFG. 
    In SLFG, $q$, $R_q$, $E_q$, and their class contexts are fed into a fine-tuned language model for logical form generation (\textcircled{5}).}
    \label{fig:framwork}
\end{figure*}

A graph structured-KB $\mathcal{G}$ is composed of a set of relational facts $\{ \langle s, r, o \rangle |s \in \mathcal{E}, r \in \mathcal{R}, o \in \mathcal{E} \cup \mathcal{L}\}$ and an ontology $\{ \langle c_d, r, c_r \rangle |c_d, c_r \in \mathcal{C}, r \in \mathcal{R} \}$.
Here, $\mathcal{E}$ denotes a set of entities, $\mathcal{R}$  a set of relations, and $\mathcal{L}$ a set of literals, e.g., textual labels or numerical values. In a relational fact $\langle s, r, o \rangle$, $s \in \mathcal{E}$ is the 
\textit{subject}, $o \in \mathcal{E} \cup \mathcal{L}$ is the \textit{object}, and $r \in \mathcal{R}$ represents the relation between the two.

The ontology defines the rules governing the composition of relational facts within $\mathcal{G}$:  $\mathcal{C}$ denotes a set of classes, each of which defines a set of entities (or literals) sharing common properties (relations). Note that an entity can belong to multiple classes. 
In an ontology triple $\langle c_d, r, c_r \rangle$, $c_d$ is the \textit{domain class}, i.e., the class of  subject entities that satisfy relation $r$; 
$c_r$ is the \textit{range class}, i.e., the class of object entities or literals satisfying $r$. Each ontology triple can be instantiated a set of relational facts, with an example provided in Appendix~\ref{sec:app_preliminaries}.

% In Figure~\ref{fig:intro_example}, \textsf{<book.literary\_series, book.literary\_series.author, book.author>} is an ontology triple. An instance of it is \textsf{<Harry Potter, book.literary\_series.author, J.K. Rowling>}, where \textsf{Harry Potter} is an entity that belongs to class \textsf{book.literary\_series}. 

\paragraph{Problem Statement} Given a KB $\mathcal{G}$ and a question $q$ expressed in natural language, i.e., a sequence of word tokens, KBQA aims to find a subset (the {answer set}) $\mathcal{A} \subseteq \mathcal{E} \cup \mathcal{L}$ of elements from $\mathcal{G}$ that --- with optional application of some aggregation functions (e.g., \textsc{count}) --- answers $q$. 

\paragraph{Logical Form}
We approach the KBQA problem by translating question $q$ into a structured query that can be executed on $\mathcal{G}$ to fetch the answer set $\mathcal{A}$. Following previous works~\cite{shu_tiara_2022, gu_dont_2023, zhang_fc-kbqa_2023}, we use logical form as the structured query language, expressed in \emph{S-expression}~\cite{gu_beyond_2021}.  
S-expression offers a readable representation well-suited for KBQA. It uses set semantics where functions operate on entities or entity tuples without requiring variables~\cite{ye_rng-kbqa_2022}, with more details in Appendix~\ref{sec:app_preliminaries}.

\section{The \model\ Model}\label{sec:method}

\model\ takes a generation-based approach overall. 
It introduces two novel modules: \emph{Schema-guided Entity Retrieval} (SER) and \emph{Schema-guided Logical Form Generation} (SLFG), designed to enhance generalizability and shown in Figure~\ref{fig:framwork}. 

\model\ starts with relation retrieval, where a BERT-based relation ranking model retrieves candidate relations and entities from the KB $\mathcal{G}$ that are potentially relevant to the question $q$. 

In SER, $q$ and the top-ranked candidate relations are passed into a logical form parser (i.e. a Seq2Seq model) to generate logical form sketches that contain entity mentions while masking out relations and classes. The retrieved relations provide the most relevant---and potentially unseen---relations as additional schema context, enabling the model to identify boundaries of entity mentions more accurately as explained in Section~\ref{sec:introduction}. We then harvest these entity mentions and use them to retrieve candidate entities from $\mathcal{G}$, thereby improving the non-I.I.D. generalizability of entity retrieval. 

To further improve the accuracy of entity retrieval, we propose a combined schema-based pruning strategy to filter out unlikely candidates, as a single mention may correspond to multiple entities. The remaining entities are then ranked by a BERT-based model, which estimates the likelihood of each entity being the correct match for a mention. Leveraging relations---a type of schema item---to guide both entity mention extraction and candidate entity pruning enhances model generalizability over entities unseen at training. This in turn helps logical form generation to filter false positive matches for unseen relations or their compositions.

In SLFG, \model\ feeds $q$, the top-ranked relations and entities (corresponding to each mention), and the schema contexts, i.e., their class information, into an adapted LLM to generate the logical form  and produce answer set $\mathcal{A}$.  SLFG is novel in that it takes (1) multiple candidate entities (instead of one in existing models) for each mention and (2) the schema contexts as the input. 

By deferring \emph{entity disambiguation} to the generation stage, our approach helps mitigate error propagation that often arises from early-stage disambiguation. 
This strategy also brings challenges, as the extra candidate entities (which often share the same or similar names) introduce noise in SLFG. 

To address this and enhance generalizability to unseen compositions of KB elements, we incorporate schema context in  SLFG. The LLM is fine-tuned to generate logical forms that consist of valid KB element compositions, as connected based on  the (domain and range) classes of the relations and entities. 

As a result, the model is able to select correct compositions from a noisy KB element space, even under non-I.I.D settings.

\subsection{Relation Retrieval}\label{subsec:ger}

For relation retrieval, we follow TIARA~\cite{shu_tiara_2022} for its high accuracy. We extract a  set $R_q$ of top-$k_R$ (system parameter) relations with the highest semantic similarity to $q$. This is done by a BERT-based cross-encoder to learn the semantic similarity between $q$ and a relation $r \in \mathcal{R}$: %(recall that $\mathcal{R}$ is the set of relations of KB $\mathcal{G}$):  
\begin{equation}
\small
    \text{sim}(q,r)=\text{\large L\small INEAR}(\text{\large B\small ERT\large C\small LS}([q;r])), 
    \label{eqn:relation_retriever}
\end{equation} 
where `$;$' denotes concatenation.
This model is trained with the sentence-pair classification objective~\cite{devlin_bert_2019}, where a relevant question-relation pair has a similarity of 1, and 0 otherwise.

\subsection{Schema-guided Entity Retrieval}

\paragraph{Entity Mention Detection} Given $R_q$, we propose a schema-guided logical form sketch parser to parse $q$ into a logical form sketch $s$. Entity mentions in $q$ are extracted from $s$. 

The parser is an adapted Seq2Seq model. The model input of each training sample takes the form of ``$q$ \textless relation\textgreater \text{ } $r_1;r_2;\ldots;r_{k_R}$'' ($r_i \in R_q$, hence ``relation-guided''). 
In the ground-truth logical form corresponding to $q$, we mask the relations, classes, and literals with special tokens `\textless relation\textgreater', `\textless class\textgreater', and `\textless literal\textgreater', to form the ground-truth logical form sketch $s$. Entity IDs are also replaced by the corresponding entity names (entity mentions), to enhance the Seq2Seq model's understanding of the semantics of entities.

At model inference, from the output top-$k_L$ (system parameter) logical form sketches  (using beam search), we extract the entity mentions.

\paragraph{Candidate Entity Retrieval}
We follow previous studies~\cite{faldu_retinaqa_2024, luo_chatkbqa_2024,shu_tiara_2022} and use an entity name dictionary FACC1~\cite{gabrilovich_facc1_2013} to map extracted entity mentions to entities (i.e., their IDs in KB), although other retrieval models can be used. Since different entities may share the same name, the entity mentions may be mapped to many entities. For pruning, existing studies use  popularity scores of the  entities~\cite{shu_tiara_2022, ye_rng-kbqa_2022}. 

\begin{figure}[t]
    \centering
    \includegraphics[width=\linewidth]{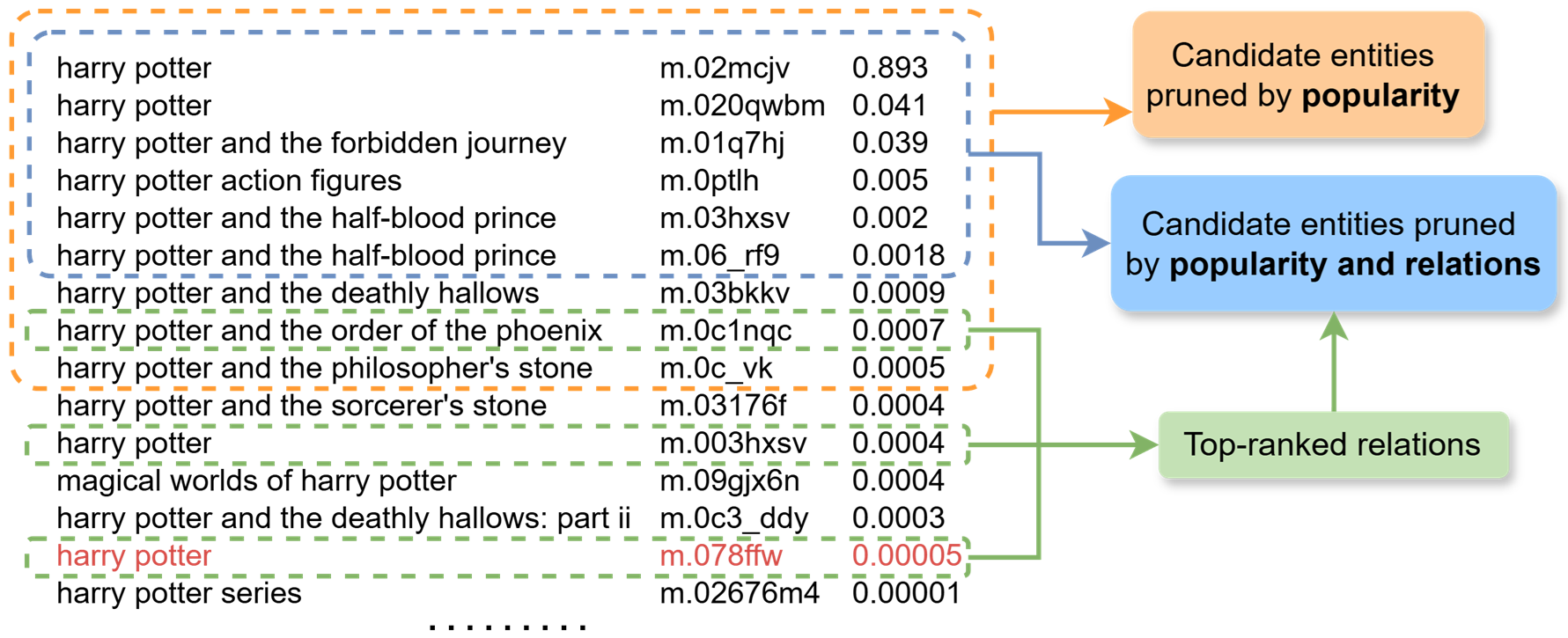}
     \caption{Candidate entity retrieval for `\textsf{Harry Potter}'. The candidate entity in red is the ground-truth.}
    \label{fig:candidate_entity_retrieval}
\end{figure}

To improve the recall, we propose a combined pruning strategy based on both popularity and relations. As Figure~\ref{fig:candidate_entity_retrieval} shows, we first select the top-$k_{E1}$ (system parameter) entities for each  mention based on popularity and then extract $k_{E2}$ (system parameter) entities from the remaining candidates that are connected to the retrieved relations $R_q$. Together, these form the candidate entity set $E_c$.

\paragraph{Entity Ranking} We follow existing works~\cite{shu_tiara_2022, ye_rng-kbqa_2022} to score and rank each candidate entity in $E_c$ by jointly encoding $q$ and the context (entity name and its linked relations) of the entity using a cross-encoder (like Eq.~\ref{eqn:relation_retriever}). %The context of a candidate entity includes . 
We select the top-$k_{E3}$ (system parameter) ranked entities for each mention as the entity set $E_q$ for each question.

\subsection{Schema-Guided Logical Form Generation}\label{subsec:rlg}

Given relations $R_q$ and entities $E_q$, we fine-tune an open-souce LLM (LLaMA3.1-8B~\cite{touvron_llama_2023} by default) to generate the final logical form.  

Before being fed into the model, each relation and entity is augmented with its class information to help the model learn their connections and generalize to unseen entities, relations, or their compositions.  The context of a relation $r$ is described by concatenating its  domain class $c_d$ and range class $c_r$, formatted as ``[D] $c_d$ [N] $r$ [R] $c_r$''. For an entity $e$, its context is described by its ID (``$id_e$''), name (``$name_e$''), and the intersection between its set of classes $C_e$ and the set of all domain and range classes $C_R$ of all relations in $R_q$, formatted as ``[ID] $id_e$ [N] $name_e$ [C] class($C_e\cap C_R$)''.

As Figure~\ref{fig:framwork} shows, we construct the input to the logical form generation model by concatenating $q$ with the context of each relation in $R_q$ and the context of each entity in $E_q$. The model is fine-tuned with a cross-entropy-based objective:
\begin{equation}
\small
\mathcal{L}_{\text{generator}}=-\sum_{t=1}^n \log p\left(l_t \mid l_{<t}, q, K_q\right),
\end{equation}
where $l$ denotes a logical form of $n$ tokens and $l_t$ is its $t$-th token, and $K_q$ is the retrieved knowledge (i.e., relations and entities with class contexts) for $q$. At inference, the model runs beam search to generate top-$k_O$ logical forms -- the executable one with the highest score is  the output. See Appendix~\ref{app:prompt} for a prompt example used for inference. 

It is possible that no generated logical forms are executable. In this case, we fall back to following~\citet{shu_tiara_2022} and~\citet{ye_rng-kbqa_2022} and retrieve candidate logical forms in two stages: enumeration and ranking. During enumeration, we traverse the KB starting from the retrieved entities. Due to the exponential growth in the candidate paths with each hop, we start from the top-1 entity for each mention and examine its neighborhood for up to two hops. The paths  are converted into logical forms. During ranking, a BERT-based ranker scores $q$ and each logical form $l$ (like Eq.~\ref{eqn:relation_retriever}). We train the ranker using a contrastive objective: 
\begin{equation}
\small
    \mathcal{L}=-\frac{\text{exp}(\text{sim}(q, l^*))}{\text{exp}(\text{sim}(q, l^*))+\sum_{l \in C_l \wedge l \neq l^*} \text{exp}(\text{sim}(q, l))},
\end{equation}
where $l^*$ is the ground-truth logical form and $C_l$ is the set of enumerated logical forms. The top-ranked, executable
logical form is returned.

\begin{table*}[!ht]
\centering
\small
\begin{tabular}{
>{\centering\arraybackslash}m{0.1\linewidth}
>{}p{0.26\linewidth}
>{\centering\arraybackslash}m{0.04\linewidth}
>{\centering\arraybackslash}m{0.04\linewidth} 
>{\centering\arraybackslash}m{0.04\linewidth} 
>{\centering\arraybackslash}m{0.04\linewidth} 
>{\centering\arraybackslash}m{0.04\linewidth} 
>{\centering\arraybackslash}m{0.04\linewidth} 
>{\centering\arraybackslash}m{0.04\linewidth} 
>{\centering\arraybackslash}m{0.04\linewidth} }
\toprule
\multicolumn{1}{l}{\textbf{}} & \textbf{} & \multicolumn{2}{c}{\textbf{Overall}} & \multicolumn{2}{c}{\textbf{I.I.D.}} & \multicolumn{2}{c}{\textbf{Compositional}} & \multicolumn{2}{c}{\textbf{Zero-shot}} \\ 
\cmidrule(lr){3-4} \cmidrule(lr){5-6} \cmidrule(lr){7-8} \cmidrule(lr){9-10}
\multicolumn{1}{l}{} & \rule{0pt}{10pt}\textbf{Model} & \textbf{EM} & \textbf{F1} & \textbf{EM} & \textbf{F1} & \textbf{EM} & \textbf{F1} & \centering \textbf{EM} & \textbf{F1} \\ \midrule
\multirow{7}{*}{\begin{tabular}[c]{@{}c@{}} SP-based \\(SFT) \\ \end{tabular}} & RnG-KBQA & 68.8 & 74.4 & 86.2 & 89.0 & 63.8 & 71.2 & 63.0 & 69.2 \\
 & TIARA & 73.0 & 78.5 & 87.8 & 90.6 & 69.2 & 76.5 & 68.0 & 73.9 \\
 & Decaf & 68.4 & 78.7 & 84.8 & 89.9 & 73.4 & \underline{81.8} & 58.6 & 72.3 \\
 & Pangu (T5-3B) & 75.4 & \underline{81.7} & 84.4 & 88.8 & \underline{74.6} & 81.5 & 71.6 & \underline{78.5} \\
 & FC-KBQA & 73.2 & 78.7 & \underline{88.5} & \underline{91.2} & 70.0 & 76.7 & 67.6 & 74.0 \\
 & TIARA+GAIN & \underline{76.3} & 81.5 & \underline{88.5} & \underline{91.2} & 73.7 & 80.0 & \underline{71.8} & 77.8 \\
 & RetinaQA & 74.1 & 79.5 & - & - & 71.9 & 78.9 & 68.8 & 74.7 \\ \midrule
\multirow{3}{*}{\begin{tabular}[c]{@{}c@{}} SP-based \\(Few-shot) \\ \end{tabular}} 
 & KB-Binder (6)-R & 53.2 & 58.5 & 72.5 & 77.4 & 51.8 & 58.3 & 45.0 & 49.9 \\
 & Pangu (Codex) & 56.4 & 65.0 & 67.5 & 73.7 & 58.2 & 64.9 & 50.7 & 61.1 \\
 & FlexKBQA & 62.8 & 69.4 & 71.3 & 75.8 & 59.1 & 65.4 & 60.6 & 68.3 \\ \midrule
\multirow{2}{*}{\centering \makecell{\textbf{Ours} \\ (SFT)}} &\textbf{\model} & \textbf{79.1} & \textbf{84.4} & \textbf{88.6} & \textbf{91.6} & \textbf{77.9} & \textbf{85.1} & \textbf{75.4} & \textbf{80.8} \\
 &\hspace{3pt} - Improvement & +3.6\% & +3.3\% & +0.1\% & +0.4\% & +4.4\% & +4.0\% & +5.0\% & +2.9\% \\ 
 \bottomrule 
\end{tabular}
\caption{\emph{Hidden} test results (\%) on GrailQA (best results are in boldface; best baseline results are underlined; ``SFT'' means supervised fine-tuning; ``few-shot'' means few-shot in-context learning).}
\label{tab:grailqa}
\end{table*}

\section{Experiments}\label{sec:experiment}
We run experiments to answer:
%\textbf{Q1}:~How does \model\ compare with SOTA models in their accuracy for the KBQA task? 
\textbf{Q1}:~How does \model\ improve generalizability compared with SOTA models?
%\textbf{Q2}:~How do model components impact the accuracy of \model? 
\textbf{Q2}:~How do model components contribute to  generalizability?
%\textbf{Q3}:~How do our techniques generalize to other KBQA models? 
\textbf{Q3}:~How can our techniques enhance existing models?

\subsection{Experimental Setup}\label{sec:experiment_setup}

\paragraph{Datasets}
Following SOTA competitors~\citep{shu_tiara_2022, gu_dont_2023, zhang_fc-kbqa_2023}, we use three benchmark datasets built upon Freebase.

\textbf{GrailQA}~\citep{gu_beyond_2021} is a dataset for evaluating the generalizability of KBQA models. It has 64,331 questions with target S-expressions, including complex questions requiring up to 4-hop reasoning over the KG. %, with functions including comparatives, superlatives, and counting. 
The dataset comes with training (70\%), validation (10\%), and test (20\%, hidden and only known by the leaderboard organizers) sets. In the validation and the test sets, 50\% of the questions include KB elements that are unseen in the training set (\textbf{zero-shot} generalization tests), 25\% consist of unseen compositions of KB elements seen in the training set (\textbf{compositional} generalization tests), and the remaining 25\% are randomly sampled from the training set (\textbf{I.I.D.} tests).

WebQuestionsSP (\textbf{WebQSP})~\citep{yih_value_2016} is a dataset for the \textbf{I.I.D.} setting. While our focus is on non-I.I.D. settings, we include results on this dataset to show the general applicability of \model. WebQSP contains 4,937 questions collected from Google query logs, including 3,098 questions for training and 1,639 for testing, each annotated with a target SPARQL query. We follow GMT-KBQA~\cite{hu_logical_2022} and TIARA~\cite{shu_tiara_2022} to separate 200 questions from the training questions to form the validation set.
% More details of WebQSP are included in Appendix~\ref{app:WebQSP}.

ComplexWebQuestions (\textbf{CWQ})~\citep{talmor_cwq_2018} is a challenging \textbf{I.I.D.} dataset for testing complex KBQA. It consists of 34,689 questions requiring up to 4-hop reasoning, constructed by extending the WebQSP dataset to questions with higher-hop complexity. For our experiments, we adopt the official split of CWQ, where 80\% of the questions are allocated for training, 10\% for validation, and the remaining 10\% for testing.

\begin{table}[t]
\centering
\small
\resizebox{\columnwidth}{!}{
\begin{tabular}{clcc}
\toprule
 & \textbf{Model} & \textbf{WebQSP} & \textbf{CWQ}\\ \midrule
\multirow{3}{*}{\begin{tabular}[c]{@{}c@{}}IR-based\\ \end{tabular}}  
& SR+NSM   & 69.5 & 47.1 \\
& UniKGQA   & 75.1  & 49.4  \\
& EPR+NSM  & 71.2  & 61.2 \\
 \midrule
\multirow{4}{*}{\begin{tabular}[c]{@{}c@{}}LLM-based\\(SFT)\\ \end{tabular}} 
& Pangu (ACL 2023) & 79.6 & - \\
& RoG  & 70.8 & 56.2\\
& ChatKBQA  & 79.8 & \underline{77.8}\\ 
& TFS-KBQA & \underline{79.9} & 63.6\\
\midrule
\multirow{5}{*}{\begin{tabular}[c]{@{}c@{}}LLM-based\\(Few-shot)\\ \end{tabular}} 
& KB-Binder (6)-R   & 53.2 & -\\
& Pangu (Codex)  & 54.5 & - \\
& FlexKBQA  & 60.6 & - \\
& ToG (GPT-4-turbo)\textsuperscript{†} & 72.3 &  56.9\\
& ICSU (SPARQL) & 72.3 & - \\
& FiDeLiS  & 78.3 & 64.3\\
\midrule
\multirow{2}{*}{\begin{tabular}[c]{@{}c@{}}\textbf{Ours} \\ (SFT)\end{tabular}} 
& \textbf{\model} & \textbf{80.3} & \textbf{78.2}\\
&\hspace{6pt} - Improvement & +0.5\% & +0.5\% \\
\bottomrule
\end{tabular}
}
\caption{F1 results (\%) on WebQSP (I.I.D.) and CWQ (I.I.D.). \textsuperscript{†}\ denotes the results we reproduced.}
\label{tab:webqsp}
\end{table}

\paragraph{Competitors}
We compare with both IR-based and SP-based methods including the SOTA models. 

On GrailQA, we compare with models that top the leaderboard\footnote{https://dki-lab.github.io/GrailQA/}, including \textbf{RnG-KBQA} (ACL 2021)~\citep{ye_rng-kbqa_2022}, \textbf{TIARA} (EMNLP 2022)~\citep{shu_tiara_2022}, \textbf{DecAF} (ICLR 2023)~\citep{yu_decaf_2023}, \textbf{Pangu} (ACL 2023, SOTA before \model, as of 19th May, 2025)~\citep{gu_dont_2023}, \textbf{FC-KBQA} (ACL 2023)~\citep{zhang_fc-kbqa_2023}, \textbf{TIARA+GAIN} (EACL 2024)~\citep{shu_data_2024}, and \textbf{RetinaQA} (ACL 2024)~\citep{faldu_retinaqa_2024}. We also compare with few-shot LLM-based (training-free) methods: KB-BINDER (6)-R (ACL 2023)~\citep{li_few-shot_2023}, Pangu~\citep{gu_dont_2023}, and FlexKBQA (AAAI 2024)~\citep{li_flexkbqa_2024}. These models are SP-based. On the non-I.I.D. GrailQA, IR-based methods are uncompetitive and excluded.

On WebQSP and CWQ, we compare with IR-based models \textbf{SR+NSM} (ACL 2022)~\citep{zhang_subgraph_2022}, \textbf{UNIKGQA} (ICLR 2023)~\citep{jiang_unikgqa_2023}, and
\textbf{EPR+NSM} (WWW 2024)~\citep{ding_enhancing_2024}, plus LLM-based supervised fine-tuning (SFT) models including \textbf{ChatKBQA} (ACL 2024)({SOTA})~\citep{luo_chatkbqa_2024}, \textbf{TFS-KBQA} (LREC-COLING 2024, {SOTA})~\citep{wang_no_2024},  and \textbf{RoG} (ICLR 2023)~\citep{luo_ReasoningOnGraph_2024}. We also compare with few-shot LLM-based methods: KB-Binder (6)-R, Pangu (Codex), FlexKBQA, \textbf{ToG} (ICLR 2024)~\citep{sun_ThinkonGraph_2024}, \textbf{ICSU} (KSEM 2024)~\citep{liu_icsu_2024}, and \textbf{FiDeLiS} (ACL 2025)~\citep{sui_FiDeLiSFaithfulReasoning_2025}. Appendix~\ref{sec:app_baselines} details these models. The baseline results are collected from their papers or the GrailQA leaderboard (when available), with ToG results reproduced due to missing F1 scores on WebQSP and CWQ.

\begin{table*}[t]
\small
\centering
\resizebox{\textwidth}{!}{
\begin{tabular}{
>{}p{0.3\linewidth}
>{\centering\arraybackslash}m{0.05\linewidth}
>{\centering\arraybackslash}m{0.05\linewidth} 
>{\centering\arraybackslash}m{0.05\linewidth} 
>{\centering\arraybackslash}m{0.05\linewidth} 
>{\centering\arraybackslash}m{0.05\linewidth} 
>{\centering\arraybackslash}m{0.05\linewidth} 
>{\centering\arraybackslash}m{0.05\linewidth} 
>{\centering\arraybackslash}m{0.05\linewidth}
c}
\toprule
& \multicolumn{8}{c}{\textbf{GrailQA}} & \multicolumn{1}{c}{\textbf{WebQSP}} \\
\cmidrule(lr){2-9} \cmidrule(lr){10-10}
& \multicolumn{2}{c}{\textbf{Overall}} 
& \multicolumn{2}{c}{\textbf{I.I.D.}} 
& \multicolumn{2}{c}{\textbf{Compositional}} 
& \multicolumn{2}{c}{\textbf{Zero-shot}} 
& \textbf{Overall} \\ 
\cmidrule(lr){2-3} \cmidrule(lr){4-5} \cmidrule(lr){6-7} \cmidrule(lr){8-9} \cmidrule(lr){10-10}
\textbf{Model}
& \textbf{EM} & \textbf{F1} & \textbf{EM} & \textbf{F1} & \textbf{EM} & \textbf{F1} & \textbf{EM}  & \textbf{F1} & \textbf{F1} \\ \midrule
\textbf{\model} & \textbf{85.1} & \textbf{88.5} & \textbf{93.1} & \textbf{94.6} & \textbf{78.4} & \textbf{83.6} & \textbf{84.4} & \textbf{87.9} & \textbf{80.3} \\
\hdashline
  \rule{0pt}{10pt}\hspace{6pt} w/o SER & 81.0 & 84.9 & 90.1 & 91.9 & 73.9 & 79.6 & 80.0 & 84.0 & 78.1 \\ 
\hspace{6pt} w/o DED & 84.3 & 87.8 & 92.6 & 94.0 & 77.1 & 82.4 & 83.7 & 87.2 & 78.2 \\ 
\hspace{6pt} w/o SC & 76.6 & 79.2 & 91.7 & 92.9 & 72.3 & 77.4 & 71.7 & 73.9 & 77.1 \\
\hspace{6pt} w/o Fallback LF & 81.8 & 84.6 & 92.8 & 94.1 & 77.3 & 81.8 & 78.7 & 81.5 & 78.6 \\
\bottomrule
\end{tabular}
}
\caption{Ablation study results on the validation set of GrailQA and the test set of WebQSP.}
\label{tab:ablation}
\end{table*}

\paragraph{Implementation Details}
All our experiments are run on a machine with an NVDIA A100 GPU and 120 GB of RAM. 
For each dataset, a T5-base model is fine-tuned for 5 epochs as our logical form sketch parser. We fine-tune a LLaMA3.1-8B with LoRA~\cite{hu_lora_2021} for 5 epochs on GrailQA and 20 epochs on WebQSP and CWQ to serve as the logical form generator. Our system parameters are selected empirically. There are only a small number of parameters to consider. As shown in the parameter study in Appendix~\ref{app:paramater_study}, our model performance shows stable patterns against the choice of parameter values. The parameter values do not take excessive fine-tuning. More implementation details are in Appendix~\ref{app:implemention_details}.

\paragraph{Evaluation Metrics}
On GrailQA, we report the exact match (\textbf{EM}) and \textbf{F1} scores, following the leaderboard. EM counts the percentage of test samples where the model generated logical form (an S-expression) that is semantically equivalent to the ground truth. F1  measures the answer set correctness, i.e., the F1 score of each answer set, average over all test samples. On WebQSP and CWQ, we report the F1 score as there are no ground-truth S-expressions. In addition, some baselines also report \textbf{Hit}, which measures whether at least one correct answer appears in the predicted answer set. Additional Hit results are reported in Appendix~\ref{app:hit_result}.

\subsection{Overall Results (Q1)}
% Tables~\ref{tab:grailqa} and~\ref{tab:webqsp} show the overall comparison of \model\ with the baseline models for GrailQA and WebQSP, respectively. \model\ shows the best results across both datasets. 

Table~\ref{tab:grailqa} reports the overall comparison of \model\ with the baseline models on GrailQA, while Table~\ref{tab:webqsp} presents the corresponding results on WebQSP and CWQ. \model\ shows the best results across all three datasets.

\paragraph{Results on GrailQA}
On the overall hidden test set of GrailQA, \model\ outperforms the best baseline Pangu by 4.9\% and 3.3\% in EM and F1 scores. While the performance improvement in the I.I.D. setting is smaller, \model\ achieves substantial gains in non-I.I.D scenarios. For example, under the compositional generalization setting, it increases EM by 4.4\% and F1 by 4.0\% over the best baseline models. Similar performance gaps are observed under the zero-shot setting, i.e., 5.0\% in EM and 2.9\% in F1. 
Notably, under non-I.I.D. settings, the improvement in EM is consistently larger than that in F1, indicating that \model\ is more capable of generating logical forms that precisely match both the questions and the KB schema, thanks to the class information that indicate the connections between the KB elements. 

Most few-shot LLM-based competitors are generally not very competitive, especially under the non-I.I.D. settings and on complex questions (CWQ). A key reason is that a limited set of examples fails to sufficiently represent the diversity of KB elements necessary to address these challenges.

\paragraph{Results on WebQSP and CWQ}
On WebQSP and CWQ, which both have I.I.D. test sets, the performance gap of different models is closer. Even in these cases, \model\ still performs the best, showing its applicability. Comparing with TFS-KBQA (SOTA on WebQSP) and ChatKBQA (SOTA on CWQ), \model\ improves the F1 score by 0.5\% on both datasets. Among IR-based methods, UniKGQA (SOTA on WebQSP) and EPR+NSM (SOTA on CWQ) still perform much worse than \model. The lower performance of IR-based methods is consistent with existing results~\citep{gu_knowledge_2022}.

\begin{table*}[t]
\small
\centering
\begin{tabular}{
>{\raggedright\arraybackslash}p{0.30\linewidth}
>{\centering\arraybackslash}m{0.05\linewidth}
>{\centering\arraybackslash}m{0.05\linewidth}
>{\centering\arraybackslash}m{0.05\linewidth}
>{\centering\arraybackslash}m{0.05\linewidth}
>{\centering\arraybackslash}m{0.05\linewidth}
>{\centering\arraybackslash}m{0.05\linewidth}
>{\centering\arraybackslash}m{0.05\linewidth}
>{\centering\arraybackslash}m{0.05\linewidth}}
\toprule
& \multicolumn{2}{c}{\textbf{Overall}}
& \multicolumn{2}{c}{\textbf{I.I.D.}}
& \multicolumn{2}{c}{\textbf{Compositional}}
& \multicolumn{2}{c}{\textbf{Zero-shot}} \\
\cmidrule(lr){2-3} \cmidrule(lr){4-5} \cmidrule(lr){6-7} \cmidrule(lr){8-9}
\textbf{Model} & \textbf{EM} & \textbf{F1} & \textbf{EM} & \textbf{F1} & \textbf{EM} & \textbf{F1} & \textbf{EM} & \textbf{F1} \\
\midrule
TIARA (T5-base)     & 75.3 & 81.9 & 88.4 & 91.2 & 66.4 & 74.8 & 73.3 & 80.7 \\
\hspace{6pt} w SER  & 79.5 & 84.3 & 90.3 & 92.3 & 71.2 & 78.1 & 78.3 & 83.3 \\
\hspace{6pt} w DED \& SC & 79.9 & 85.6 & 88.6 & 92.3 & 72.7 & 79.8 & 79.0 & 85.0 \\
\midrule
\textbf{SG-KBQA}        & \textbf{85.1} & \textbf{88.5} & \textbf{93.1} & \textbf{94.6} & \textbf{78.4} & \textbf{83.6} & \textbf{84.4} & \textbf{87.9} \\
\hspace{6pt} w T5-base  & 80.6 & 84.9 & 89.9 & 92.6 & 73.8 & 81.0 & 79.4 & 83.3 \\
\bottomrule
\end{tabular}
\caption{Module applicability results on the validation set of GrailQA.}
\label{tab:exp_applicability}
\end{table*}

\subsection{Ablation Study (Q2)}

% \begin{table}[t]
% \centering
% \resizebox{\columnwidth}{!}{
% \begin{tabular}{lccccc}
% \toprule
% \multirow{2}{*}{\textbf{Model}} & \multicolumn{4}{c}{\textbf{GrailQA}} & \textbf{WebQSP} \\ 
% \cmidrule(lr){2-5} \cmidrule(lr){6-6}
% & \textbf{Overall} & \textbf{I.I.D.} & \textbf{Comp.} & \textbf{Zero.} & \textbf{Overall} \\ 
% \midrule
% \textbf{\model} & \textbf{88.5} & \textbf{94.6} & \textbf{83.6} & \textbf{87.9} & \textbf{80.3} \\
% \hspace{5pt}w/o SER   & 84.9 & 91.9 & 79.6 & 84.0 & 78.1 \\
% \hspace{5pt}w/o DED   & 87.8 & 94.0 & 82.4 & 87.2 & 78.2 \\
% \hspace{5pt}w/o SC    & 79.2 & 92.9 & 77.4 & 73.9 & 77.1 \\
% \bottomrule
% \end{tabular}
% }
% \caption{Ablation study results (F1 score) on the validation set of GrailQA and the test set of WebQSP.}
% \label{tab:ablation}
% \end{table}

Next, we run an ablation study with the following variants of \model: \textbf{w/o~SER} replaces our schema-guided entity retrieval with the entity linking results from TIARA~\citep{shu_tiara_2022} which is commonly used in the baselines~\citep{gu_dont_2023, faldu_retinaqa_2024}; \textbf{w/o~DED} uses the top-1 candidate entity for each entity mention without deferring entity disambiguation; \textbf{w/o~SC} omits schema contexts (classes) from logical form generation; \textbf{w/o Fallback LF} removes the fall back logical form generation mechanism from \model. Table~\ref{tab:ablation} shows the results on the validation set of GrailQA and the test set of WebQSP.

\paragraph{Schema-guided Entity Retrieval} SER improves the F1 score by 3.7 points and the EM score by 4.1 points on GrailQA (i.e., \model\ vs. \model w/o SER for overall results), while achieving a 1.9-point increase in the F1 score on WebQSP. It yields an improvement of at least 3.9 F1 points and 4.5 EM points on the compositional and zero-shot test sets, which exceeds the gains observed on the I.I.D. test set (2.5 F1 points and 3.0 EM points). This indicates that incorporating retrieved relations helps the model more accurately identify them and entity mention boundaries in questions involving unseen relations. Since the entity retrieval results directly serve as input to the generation stage, this improvement further enhances the model’s overall generalization performance under non-I.I.D. settings. More discussion on entity retrieval results and a case study are provided in Appendix~\ref{app:retrieval_performance} and Appendix~\ref{app:case_study}, respectively.

\paragraph{Schema-guided Logical Form Generation} \model\ w/o DED negatively impacts the F1 scores on both GrailQA and WebQSP, confirming that our DED strategy effectively mitigate error propagation between the retrieval and generation stages. Meanwhile, \model w/o SC (with deferred entity disambiguation but no class information) has the most significant drops in F1 under the compositional (7.2) and zero-shot (14.0) tests. This highlights the contribution of class information in enabling the model to understand the connections among retrieved KB elements, thereby facilitating the generation of correct logical forms. A case study illustrating the schema-guided logical form generation module is provided in Appendix~\ref{app:case_study}.

\begin{figure}[t]
    \centering
    \includegraphics[width=\linewidth]{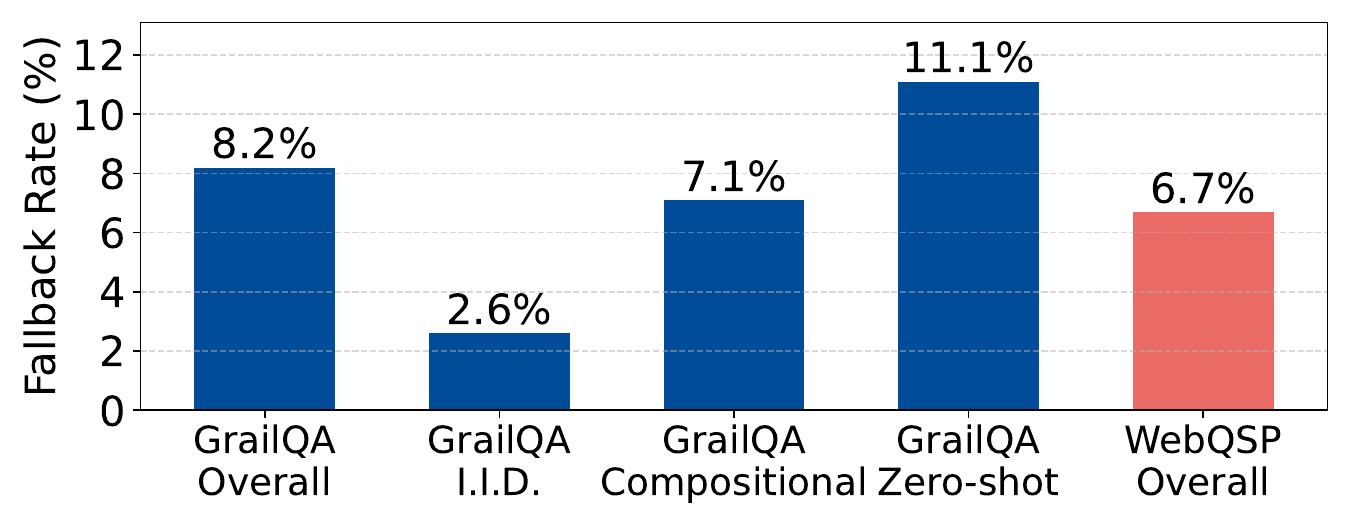}
    \caption{Proportion of questions resolved by the fallback mechanism on the validation set of GrailQA and the test set of WebQSP.}
    \label{fig:fallback}
\end{figure}

\paragraph{Fallback Mechanism} \model w/o Fallback LF exhibits slightly lower performance, with F1 decreasing by 0.9 and EM by 3.3 on GrailQA, and F1 decreasing by 1.7 on WebQSP. We note that this fallback mechanism is \emph{not} the reason why \model\ outperforms the baseline models. TIARA also uses this fallback mechanism, while RetinaQA uses the top executable logical form from the fallback mechanism as one of the options to be selected by its discriminator to determine the final logical form output. As shown in Table~\ref{tab:ablation}, \model still produces competitive results without fallback on both I.I.D and non-I.I.D settings. Figure~\ref{fig:fallback} further reports the proportion of questions where the fallback mechanism was triggered, which is at most 11.1\% of the test questions. This demonstrates that our SG-KBQA model can generate executable logical forms under non-I.I.D. settings in most cases without relying on fallback.

\subsection{Module Applicability (Q3)} \label{sec:applicability}

Our entity retrieval module \textbf{SER} and logical form generation module \textbf{DED \& SC} can be applied to existing  models to improve their generalizability under non-I.I.D. settings. We showcase such applicability with TIARA. As shown in Table~\ref{tab:exp_applicability}, replacing TIARA's entity retrieval module with ours (TIARA w SER) helps boost the EM and F1 scores by 4.2 and 2.4 points overall, comparing against the original TIARA model. This improvement is primarily from the tests with KB elements or compositions that are unseen at training, as evidenced by the larger performance gains on the compositional and zero-shot tests, i.e., 3.3 and 2.6 points in F1, respectively. Similar patterns are observed for TIARA w DED \& SC that replaces TIARA's logical form generation module with ours. These results demonstrate that our modules can enhance the retrieval and generation steps of other compatible models, especially under non-I.I.D. settings. 

Table~\ref{tab:exp_applicability} further reports EM and F1 scores of \model\ when replacing LLaMA3.1-8B with \textbf{T5-base} (which is used by TIARA) for logical form generation. We see that, even with the same T5-base model for the logical form generator, \model\ outperforms TIARA by 5.3 points 3.0 points in the EM and F1 scores for the overall tests. \emph{This further confirms that the performance gains come from the incorporation of the class contexts instead of a more advanced backbone model.} 

We also have results on parameter impact, model running time, a case study, and error analyses. They are documented in Appendices~\ref{app:paramater_study} to~\ref{app:error_analysis}.

% \begin{table}[t]
% \centering
% \resizebox{\columnwidth}{!}{
% \begin{tabular}{lcccc}
% \toprule
% \textbf{Model}                & \textbf{Overall} & \textbf{I.I.D.}& \textbf{Comp.} & \textbf{Zero.} \\ \midrule
% TIARA (T5-base)   & 81.9   & 91.2  & 74.8  & 80.7  \\ 
% \hspace{3pt} w SER           & 84.3   & 92.3  & 78.1  & 83.3  \\
% \hspace{3pt} w DED \& SC     & 85.6   & 92.3  & 79.8  & 85.0  \\
% \midrule
% \textbf{\model}            & \textbf{88.5}   & \textbf{94.6}  & \textbf{83.6}  & \textbf{87.9}  \\
% \hspace{3pt} w T5-base       & 84.9   & 92.6  & 81.0  & 83.3  \\
%  \bottomrule
% \end{tabular}
% }
% \caption{Module applicability results (F1 score) on the validation set of GrailQA. EM scores are in Appendix~\ref{app:applicability}.}\label{tab:exp_applicability}
% \end{table}

\section{Conclusion}\label{sec:conclusion}

We proposed \model for KBQA. Our core innovations include: (1) using relation to guide entity retrieval; (2) deferring entity disambiguation to the logical form generation stage; and (3) enriching logical form generation with schema (class) contexts indicate KB element connections. Together, we achieve a model that tops the leaderboard of a popular non-I.I.D. dataset GrailQA, outperforming SOTA models by 4.0\%, 2.9\%, and 3.3\% in F1 under compositional generalization, zero-shot generalization, and overall test settings, respectively. Our model also performs well in the I.I.D. setting, outperforming SOTA models on WebQSP and CWQ.

\section*{Limitations}
Like any other supervised models, \model requires annotated samples for training which may be difficult to obtain for many domains. Exploiting LLMs to generate synthetic training data is a promising direction to address this issue. 

Also, as discussed in the error analysis in Appendix~\ref{app:error_analysis}, errors can still arise from the relation retrieval, entity retrieval, and logical form generation modules. There are rich opportunities in further strengthening these modules. 
As we start from relation extraction, the overall model accuracy relies on highly accurate relation extraction. It would be interesting to explore how well \model performs on even larger KBs with more relations. 

We further noted several recent works in this highly competitive area, e.g.,
READS~\citep{xu_LLMbasedDiscriminativeReasoning_2025} and 
MemQ~\citep{xu_MemoryaugmentedQueryReconstruction_2025}, which fine-tune LLMs for reasoning trajectory generation or step-wise tool invocation based on given topic entities. These studies represent parallel efforts to ours with a focus on I.I.D. settings. We plan to evaluate their performance in future work, particularly under settings where the topic entity is unavailable and in our non-I.I.D. scenarios.

\section*{Ethics Statement}
This work adheres to the ACL Code of Ethics and is based on publicly available datasets, used in compliance with their respective licenses. As our data contains no sensitive or personal information, we foresee no immediate risks. To promote reproducibility and further research, we also open-source our code.

\section*{Acknowledgments}
This work is in part supported by the Australian Research
Council (ARC) via Discovery Project DP240101006. Jianzhong Qi is supported by ARC Future Fellowship FT240100170.

\bibliography{references}

\clearpage

\appendix

\section{Basic Concepts}\label{sec:app_preliminaries}

\begin{figure}[h]
    \centering
    \includegraphics[width=\columnwidth]{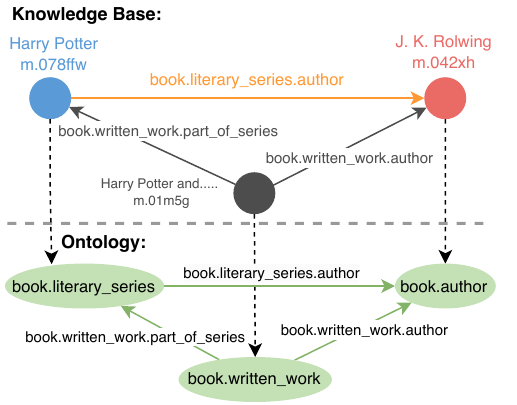}
    \caption{A subgraph of Freebase (top) and its corresponding ontology (bottom).}
    \label{fig:ontology_subgraph}
\end{figure}

% \paragraph{Ontology} As shown in Figure~\ref{fig:ontology_subgraph}, \textsf{<book.literary\_series, book.literary\_series.\allowbreak/author, book.author>} is an ontology triple. An instance of it is \textsf{<Harry Potter, book.literary\_series.author, J.K. Rowling>}, where \textsf{Harry Potter} is an entity that belongs to class \textsf{book.literary\_series}. 

\paragraph{Ontology} 
As shown in Figure~\ref{fig:ontology_subgraph}, an ontology triple example is:
\begin{quote}
\small
\textsf{book.literary\_series, \\ book.literary\_series.author, \\ book.author}
\end{quote}
An instance of it is:
\begin{quote}
\small
\textsf{Harry~Potter,\\ book.literary\_series.author,\\ J.K.~Rowling}
\end{quote}

Here, \textsf{Harry Potter} is an entity that belongs to class \textsf{book.literary\_series}; \textsf{J.K. Rowling} is an entity that belongs to class \textsf{book.author}.

\paragraph{S-expressions} S-expressions~\cite{gu_beyond_2021} use set-based semantics defined over a set of operators and operands. The operators are represented as functions. 
Each function takes a number of arguments (i.e., the operands). Both the arguments and the return values of the functions are either a set of entities or entity tuples (or tuples of an entity and a literal). The functions available in S-expressions are listed in Table~\ref{tab:logical_form_operators}, where a set of entities typically refers to a class (recall that a class is defined as a set of entities sharing common properties) or individual entities, and a binary tuple typically refers to a relation.

\begin{table*}[ht]
\small
\setlength{\tabcolsep}{2pt}
\begin{tabular}{lll}
\toprule
\multicolumn{1}{l}{\textbf{Function}}                                             & \multicolumn{1}{l}{\textbf{Return value}} & \multicolumn{1}{l}{\textbf{Description}}                                                               \\ \midrule
(\texttt{AND} $u_1$ $u_2$)                                                                       & a set of entities                    & The \texttt{AND} function returns the intersection of two sets $u_1$ and $u_2$                                               \\ \hline
(\texttt{COUNT} $u$)                                                                         & a singleton set of integers           & The \texttt{COUNT} function returns the cardinality of set $u$                                                 \\ \hline
(\texttt{R} $b$)                                                                             & a set of (entity, entity) tuples     & The \texttt{R} function reverses each binary tuple $(x, y)$ in set $b$ to $(y, x)$                                   \\ \hline
(\texttt{JOIN} $b$ $u$)                                                                        & a set of entities                    & Inner \texttt{JOIN} based on entities in set $u$ and the second element of tuples in set $b$                                    \\ \hline
(\texttt{JOIN} $b_1$ $b_2$)                                                                      & a set of (entity, entity) tuples     & Inner \texttt{JOIN} based on the first element of tuples in set $b_2$ and the second element\\
& &  of tuples in set $b_1$             \\ \hline
\begin{tabular}[c]{@{}l@{}}(\texttt{ARGMAX} $u$ $b$)\\ (\texttt{ARGMIN} $u$ $b$)\end{tabular}                & a set of entities                    & These functions return $x$ in $u$ such that $(x,y) \in b$ and $y$ is the largest / smallest                                   \\ \hline
\begin{tabular}[c]{@{}l@{}}(\texttt{LT} $b$ $n$)\\ (\texttt{LE} $b$ $n$)\\ (\texttt{GT} $b$ $n$)\\ (\texttt{GE} $b$ $n$)\end{tabular} & a set of entities                    & These functions return all $x$ such that $(x, v) \in b$ and $v$ $<$ / $\le$ / $>$ / $\ge$ $n$ \\ \bottomrule
\end{tabular}
\caption{Functions (operators) defined in S-expressions ($u$: a set of entities, $b$: a set of (entity, entity or literal) tuples, $n$: a numerical value).}\label{tab:logical_form_operators}
\end{table*}

\begin{table*}[t]
\small
\centering
\begin{tabular}{l}
\toprule
\textbf{Example Prompt} \\
\midrule

Please translate the following question into logical form using the provided relations and entities. \\ \\

\textbf{Question:} Strong lyrics is the description of which video game rating? \\ \\

\textbf{Candidate relations with their corresponding Domain [D], Name [N], Range [R]:} \\ \\
 
[D] cvg.game\_version \\ 

[N] cvg.game\_version.rating \\

[R] cvg.computer\_game\_evaluation; \\ \\

[D] cvg.computer\_game\_rating \\ 

[N] cvg.computer\_game\_rating.rating\_system \\

[R] cvg.computer\_game\_rating\_system; \\ \\

[D] cvg.computer\_game\_rating\_system \\

[N] cvg.computer\_game\_rating\_system.content\_descriptors \\

[R] cvg.computer\_game\_content\_descriptors \\ \\

...(Continue in the same manner for additional relations) \\ \\

\textbf{Candidate entities with their corresponding id [ID], Name [N], Class [C]:} \\ \\

[ID] m.042zlv3 \\

[N] Strong lyrics \\

[C] cvg.computer\_game\_content\_descriptors \\ \\

...(Continue in the same manner for additional entities) \\
\bottomrule
\end{tabular}
\caption{Example prompt to our fine-tuned LLM-based logical form generator for an input question: \textsf{Strong lyrics is the description of which video game rating?}}
\label{tab:prompt_example}
\end{table*}

\section{Prompt Example}\label{app:prompt}

We show an example prompt to our fine-tuned LLM-based logical form generator containing top-20 relations and top-2 entities per mention retrieved by our model in Table~\ref{tab:prompt_example}.

\section{Baseline Models}\label{sec:app_baselines}
The following models are tested against \model\ on the GrailQA dataset:
\begin{itemize}
    \item RnG-KBQA~\cite{ye_rng-kbqa_2022} enumerates and ranks all possible logical forms within two hops from the entities retrieved by an entity retrieval step. It  uses a Seq2Seq model to generate the target logical form based on the input question and the top-ranked candidate logical forms.
    \item TIARA~\cite{shu_tiara_2022} shares the same overall procedure with RnG-KBQA. It further retrieves entities, relations, and classes based on the input question and feeds these KB elements into the Seq2Seq model together with the question and the top-ranked candidate logical forms to generate the target logical form.

    \item TIARA+GAIN~\cite{shu_data_2024} enhances TIARA using a training data augmentation strategy. It synthesizes additional question-logical form pairs for model training to enhance the model's capability to handle more entities and relations. This is done by a 
     graph traversal to randomly sample logical forms from the KB  and a PLM to generate questions corresponding to the logical forms (i.e., the ``GAIN'' module). TIARA+GAIN is first tuned using the synthesized data and then tuned on the target dataset, for its retriever and generator modules which both use PLMs.
    
    \item Decaf~\cite{yu_decaf_2023} uses a Seq2Seq model that takes as input a question and a linearized question-specific subgraph of the KG and jointly decodes into both a  logical form and an answer candidate. The logical form is then executed, which produces a second answer candidate if successful. The final answer is determined from these two answer candidates with a scorer model. 

    \item Pangu~\cite{gu_dont_2023} formulates logical form generation as an iterative  enumeration process starting from the entities retrieved by an entity retrieval step. 
    At each iteration, partial logical forms generated so far are extended following paths in the KB to generate more and longer partial logical forms. A language model is used to select the top partial logical forms to be explored in the next iteration, under either fined-tuned models (T5-3B) or few-shot in-context learning (Codex). 
    
    \item FC-KBQA~\cite{zhang_fc-kbqa_2023} employs an intermediate module to test the connectivity between the retrieved KB elements, and it  generates the target logical form using the connected pairs of the retrieved KB elements through a Seq2Seq model.
     
    \item RetinaQA~\cite{faldu_retinaqa_2024} is a two-branch model where one follows a ranking-based approach while the other follows a sketch-filling-based logical form construction method.  It then uses a discriminator to determine the final output logical form from the two branches.We note that while we also generate logical form sketches. Such sketches are used for entity mention detection only and is not used to form the final output logical forms directly, i.e., our method is \emph{not} sketch-filling-based.
    
    \item KB-BINDER~\cite{li_few-shot_2023} uses a training-free few-shot in-context learning model based on LLMs. It generates a draft logical form by showcasing the LLM examples of questions and logical forms (from the training set) that are similar to the given test question. Subsequently, a retrieval module grounds the surface forms of the KB elements in the draft logical form to specific KB elements.

    \item FlexKBQA~\cite{li_flexkbqa_2024} considers limited training data and leverages an 
    LLM to generate additional training data. 
    It samples executable logical forms from the KB and utilizes an LLM with few-shot in-context learning to convert them into natural language questions, forming synthetic training data. These data, together with a few real-world training samples, are used to train a KBQA model. Then, the model is used to generate logical forms with more real world questions (without ground truth), which are filtered through an execution-guided module to prune the erroneous ones. The remaining logical forms and the corresponding real-world questions are used to train a new model. This process is repeated, to align the distributions of  synthetic training data and real-world questions. 

\end{itemize}

The following models are tested against \model\ on the WebQSP dataset:

\begin{itemize}

    \item Subgraph Retrieval (SR)~\cite{zhang_subgraph_2022} focuses on retrieving a KB subgraph relevant to the input question. It does not concern retrieving the exact question answer by  reasoning over the subgraph. Starting from the topic entity, it performs a top-$k$ beam search at each step to progressively expand into a subgraph, using a scorer module to score the candidate relations to be added to the subgraph next. 

    \item Evidence Pattern Retrieval (EPR)~\cite{ding_enhancing_2024} aims to extract subgraphs with fewer noise entities. It starts from the topic entities and expands by retrieving and ranking atomic (topic entity-relation or relation-relation) patterns relevant to the question. This forms a set of relation path graphs (i.e., the candidate \emph{evidence patterns}). The relation path graphs are then ranked to select the most relevant one. By further retrieving the entities on the selected relation path graph, EPR obtains the final subgraph relevant to the input question. 
     
    \item Neural State Machine (NSM)~\cite{he_improving_2021} is a reasoning model to find answers for the KBQA problem from a subgraph (e.g., retrieved by SR or EPR). It address the issue of lacking intermediate-step supervision signals when reasoning through the subgraph to reach the answer entities. This is done by training a so-called teacher model that follows a bidirectional reasoning mechanism starting from both the topic entities and the answer entities. During this process, the  ``distributions'' of entities, which represent their probabilities to lead to the answer entities (i.e., intermediate-step supervision signal), are propagated. 
    A second model, the so-called student model, learns from the teacher model to generate the entity distributions, with knowledge of the input question and the topic entities but not the answer entities. Once trained, this model can be used for KBQA answer reasoning.    

    \item UniKGQA~\cite{jiang_unikgqa_2023} integrates both retrieval and reasoning stages to enhance the accuracy of multi-hop KBQA tasks. It trains a PLM to learn the semantic relevance between every relation and the input question. The semantic relevance information is  propagated and aggregated through the KB to form the semantic relevance between the entities and the input question. The entity with the highest semantic relevance is returned as the answer.

    \item ChatKBQA~\cite{luo_chatkbqa_2024} fine-tunes an open-source LLM to map questions into draft logical forms. The  ambiguous KB items in the draft logical forms are replaced with specific KB elements by a separate retrieval module.
    
    \item TFS-KBQA~\cite{wang_no_2024} fine-tunes an LLM for more accurate logical form generation with three strategies.  The first strategy directly fine-tunes the LLM to map natural language questions into draft logical forms containing entity names instead of entity IDs. The second strategy breaks the mapping process into two steps, first to generate relevant KB elements, and then to generate draft logical forms using the KB elements. The third strategy fine-tunes the LLM to directly generate the answer to an input question. 
    After applying the three fine-tuning strategies, the LLM is used to map natural language questions into draft logical forms at model inference. A separate entity linking module is used to further map the entity names in draft logical forms into entity IDs. 

    \item RoG~\citep{luo_ReasoningOnGraph_2024} fine-tunes an LLM to generate relation paths grounded in the KB, given an input question. These generated relation paths are then used to retrieve valid reasoning paths from the KB, enabling faithful reasoning to derive the final answers.

    \item ToG~\citep{sun_ThinkonGraph_2024} performs step-by-step relation path selection using an LLM via few-shot in-context learning. At each step, based on  newly extended paths, the model prompts the LLM to either answer the question based on the reasoning paths extended so far or continue with another step of path extension.

    \item ICSU~\citep{liu_icsu_2024} retrieves questions from the training set that are similar to an input question. Pairs of the retrieved questions and the corresponding logical forms are used as examples to prompt an LLM to generate logical forms for the input question.

    \item FiDeLiS~\citep{sui_FiDeLiSFaithfulReasoning_2025} combines semantic similarity metrics with graph-based connectivity to incrementally retrieve and extend reasoning paths. It further proposes Deductive-Verification Beam Search (DVBS), which converts the question into a declarative statement and uses an LLM to verify --- by combining this statement with the current tail node of the reasoning path --- whether the current node can be directly output as the answer or the path should be further extended.

\end{itemize}

\section{Implementation Details}\label{app:implemention_details}

All our experiments are run on a machine with an NVDIA A100 GPU and 120 GB of RAM. We fine-tuned three \texttt{bert-base-uncased} models for a maximum of three epochs each, for relation retrieval, entity ranking, and fallback logical form ranking. For relation retrieval, we randomly sample 50 negative samples for each question to train the model to distinguish between relevant and irrelevant relations. 

For each dataset, a \texttt{T5-base} model is fine-tuned for 5 epochs as our logical form sketch parser, with a beam size of 3 (i.e., $k_L = 3$) for GrailQA, and 4 for WebQSP. For candidate entity retrieval, we use the same number (i.e., $k_{E1} + k_{E2}  = 10$) of candidate entities per mention as that used by the baseline models~\cite{shu_tiara_2022,ye_rng-kbqa_2022}. The retrieved candidate entities for a mention consist of entities with the top-$k_{E1}$ popularity scores and $k_{E2}$ entities connected to the top-ranked relations in $R_q$, where $k_{E1} = 1$, $k_{E2} = 9$ for GrailQA, $k_{E1} = 3$, $k_{E2} = 7$ for WebQSP. We select the top-20 (i.e., $k_R$ = 20) relations and the top-2 (i.e., $k_{E3} = 2$) entities (for each entity mention) retrieved by our model. For WebQSP, we also use the candidate entities obtained from the off-the-shelf entity linker ELQ~\cite{li_efficient_2020}. 

Finally, we fine-tune LLaMA3.1-8B with LoRA~\cite{hu_lora_2021} for logical form generation. On GrailQA, LLaMA3.1-8B is fine-tuned for 5 epochs with a learning rate of $0.0001$. On WebQSP, it is fine-tuned for 20 epochs with the same learning rate (as it is an I.I.D. dataset where more epochs are beneficial). During inference, we generate logical forms by beam search with a beam size of 10 (i.e., $K_O = 10$). The generated logical forms are executed on the KB to filter non-executable ones. If none of the logical forms are executable, we check candidate logical forms from the fallback procedures, and the result of the first executable one is returned as the answer set.

Following our baselines~\citep{shu_tiara_2022,gu_dont_2023, faldu_retinaqa_2024}, all retrieval and generation models used in our approach are trained separately on the training sets of GrailQA and WebQSP, and then evaluated on their respective validation and test sets. Specifically, for GrailQA, we follow previous works~\citep{shu_tiara_2022,gu_dont_2023,faldu_retinaqa_2024} and adopt the official data splits described in Section~\ref{sec:experiment_setup} to evaluate generalization under non-I.I.D. settings. For WebQSP, we use the same data splits (described in Appendix~\ref{sec:experiment_setup}) as in previous studies~\citep{shu_tiara_2022, gu_dont_2023, luo_chatkbqa_2024, wang_no_2024} to ensure fair comparison.

Our system parameters are selected empirically. There are only a small number of parameters to consider. As shown in the parameter study later, our model performance shows stable patterns against the choice of parameter values. The parameter values do not take excessive fine-tuning. 

\section{Results on an Additional Evaluation Metric}\label{app:hit_result}

\begin{table}[h]
\centering
\small
\begin{tabular}{lccc}
\toprule
 \textbf{Model}  & \textbf{GrailQA}    & \textbf{WebQSP}  & \textbf{CWQ} \\
\midrule
ToG & 81.4 & 82.6 & 67.6 \\
FiDeLiS & - & \textbf{84.3} & 71.4 \\
SG-KBQA (ours)    & \textbf{89.0} & 84.1 & \textbf{82.1} \\
\bottomrule
\end{tabular}
\caption{Hit results (\%) on the validation set of GrailQA and the test set of WebQSP and CWQ.}
\label{tab:hit}
\end{table}

Following recent studies~\cite{sun_ThinkonGraph_2024, sui_FiDeLiSFaithfulReasoning_2025}, we also compare the Hit results of \model\ with theirs on GrailQA, WebQSP, and CWQ. Although \model\ is not explicitly optimized for producing at least one correct answer, it still shows substantial gains on GrailQA (7.6) and CWQ (10.7) over the baselines. On WebQSP, the Hit of \model\ is only 0.2 lower than the best baselines. These results further confirm the effectiveness of \model\ in enhancing generalizability on non-I.I.D. scenarios and accuracy on complex questions.

\section{Parameter Study}\label{app:paramater_study}

We conduct a parameter study to investigate the impact of the choice of values for our system parameters. When the value of a parameter is varied, default values as mentioned in Appendix~\ref{app:implemention_details} are used for the other parameters. 

\begin{figure}[h] 
    \begin{minipage}[b]{0.5\linewidth}  
        \centering
        \includegraphics[width=\textwidth]{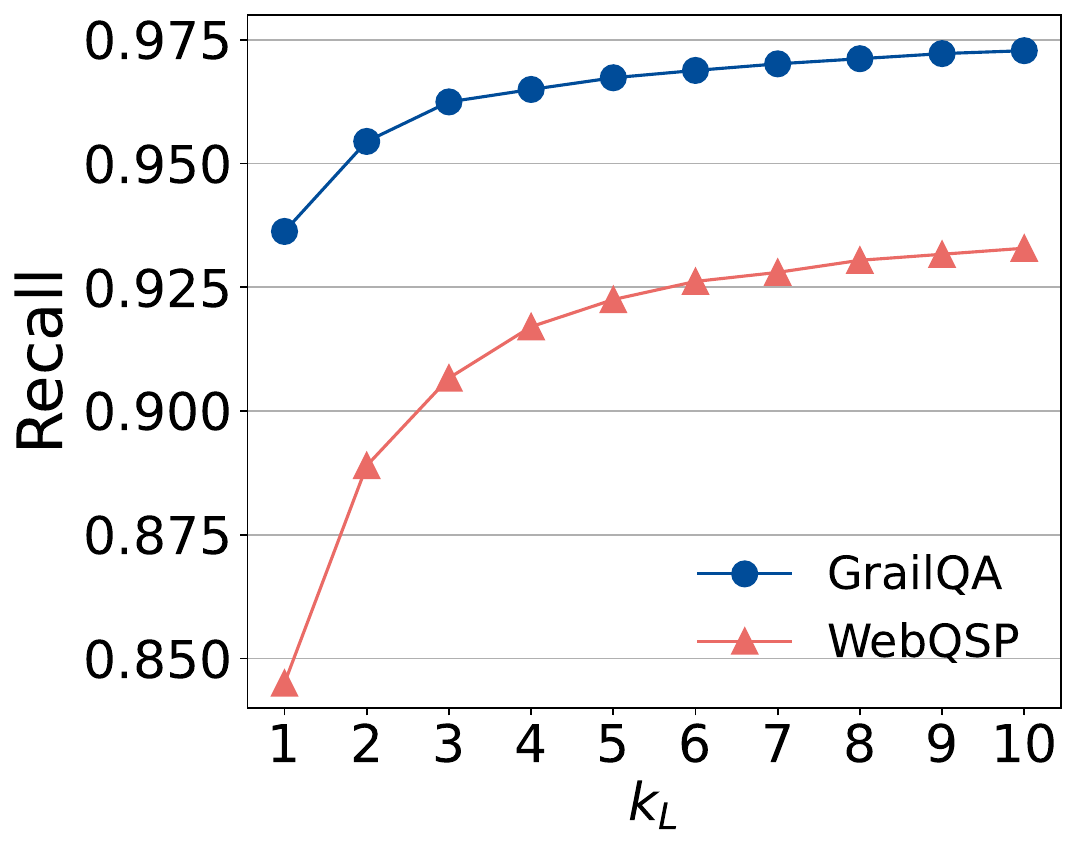} 
        \captionsetup{font=small}
        \subcaption{$k_L$}
       % \label{fig:sub1}
    \end{minipage}%
    \hfill 
    \begin{minipage}[b]{0.5\linewidth}  
        \centering
        \small
        \includegraphics[width=\textwidth]{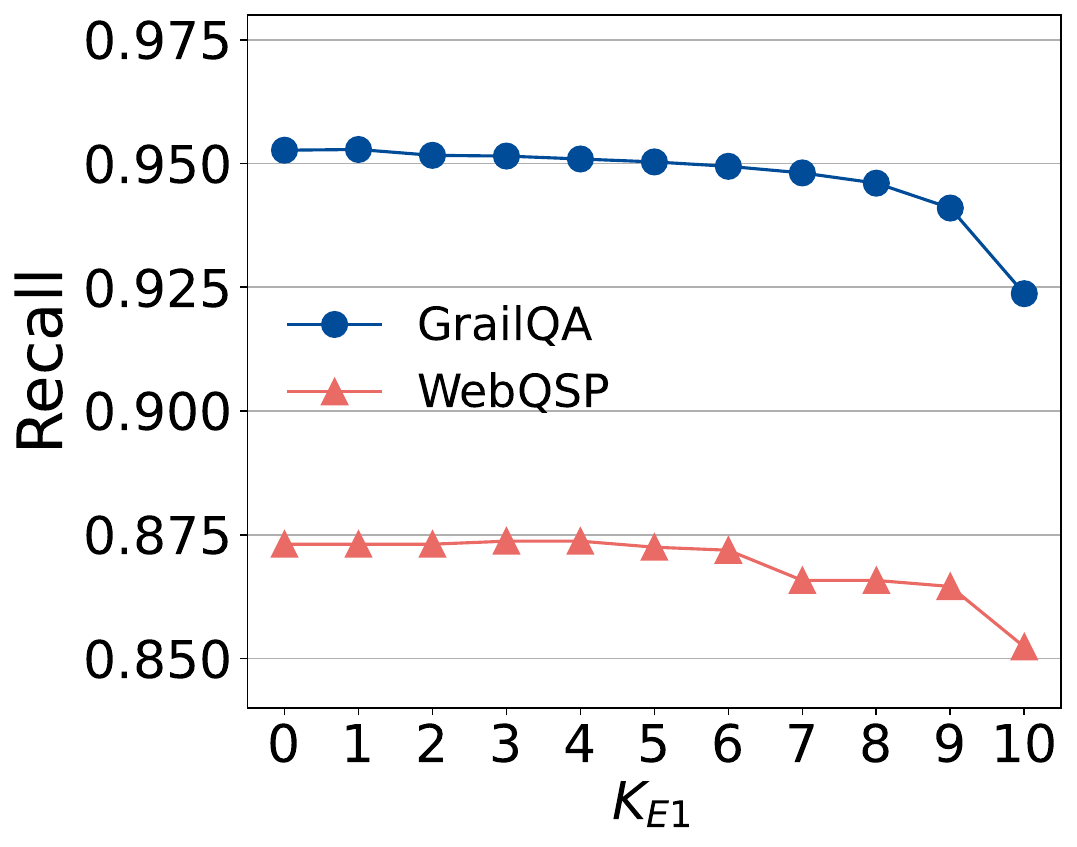}
        \captionsetup{font=small}
        \subcaption{$k_{E1}$}
        %\label{fig:sub2}
    \end{minipage}
    \caption{Impact of $k_L$ and $k_{E1}$ on the recall of candidate entity retrieval.} 
    \label{fig:kl_ke1}
\end{figure}

\begin{figure}[h] 
    \begin{minipage}[b]{0.5\linewidth}  
        \centering
        \includegraphics[width=\textwidth]{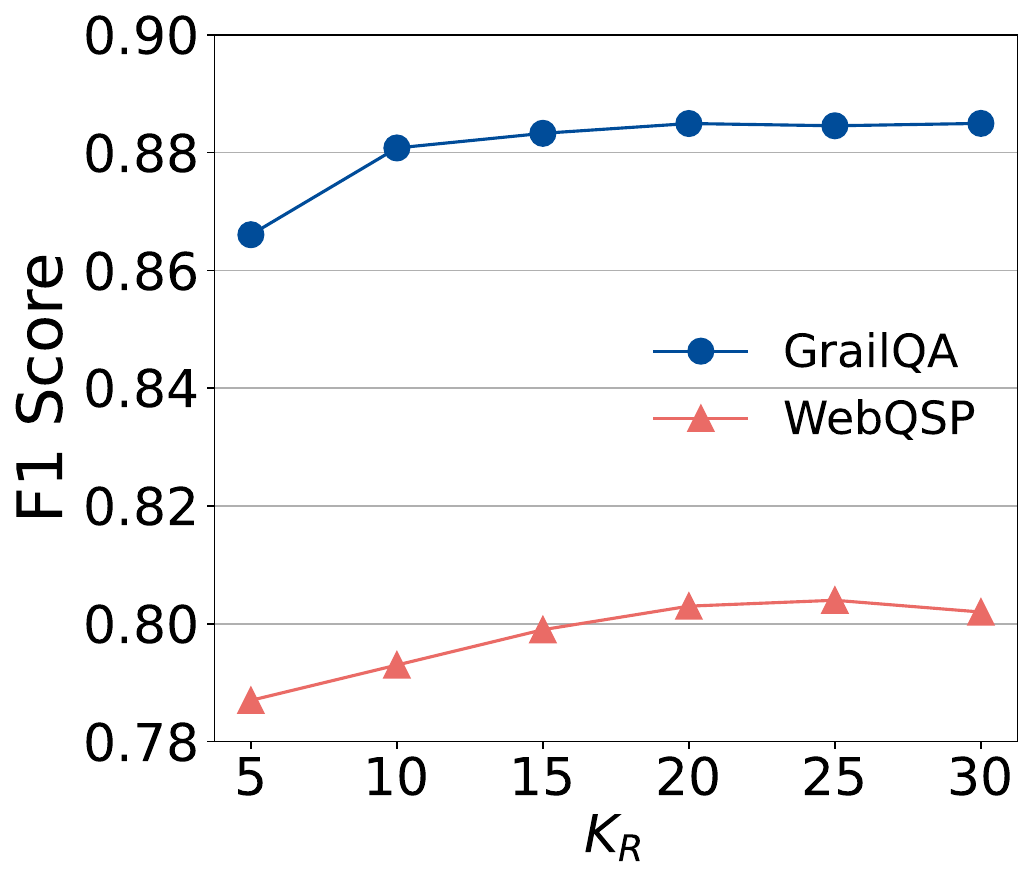} 
        \captionsetup{font=small}
        \subcaption{$k_R$}
        %\label{fig:sub1}
    \end{minipage}%
    \hfill 
    \begin{minipage}[b]{0.5\linewidth}  
        \centering
        \small
        \includegraphics[width=\textwidth]{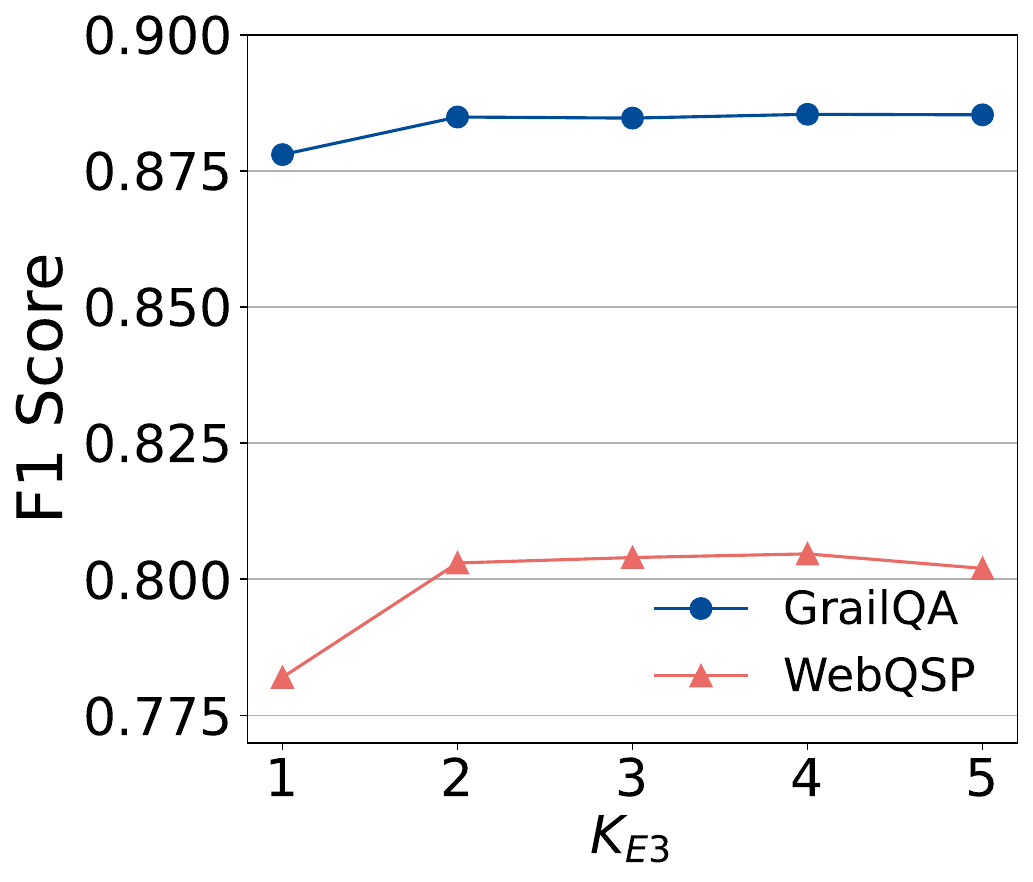}
        \captionsetup{font=small}
        \subcaption{$k_{E3}$}
        %\label{fig:kr_ke3}
    \end{minipage}
    \caption{Impact of $k_R$ and $k_{E3}$ on the overall F1 score.}
    \label{fig:kr_ke3}
\end{figure}

Figure~\ref{fig:kl_ke1} presents the impact of  $k_L$ and $k_{E1}$ on the recall of candidate entity retrieval (i.e., the average percentage of ground-truth entities returned by our candidate entity retrieval module for each test sample). Here, for the GrailQA dataset, we report the results on the overall tests (same below). 
Recall that $k_L$ means the number of logical form sketches from which entity mentions are extracted, while $k_{E1}$ refers to the number of candidate entities retrieved based on the popularity scores. 

As $k_L$ increases, the recall of candidate entity retrieval grows, which is expected. The growth diminishes gradually. This is because a small number of questions contain complex entity mentions that are difficult to handle (see error analysis in Appendix~\ref{app:error_analysis}). As $k_L$ increases, the precision of the retrieval also reduces, which brings noise into the entity retrieval results and additional computational costs. 
To strike a balance, we set $k_L = 3$ for GrailQA and  $k_L = 4$ for WebQSP. We also observe that the recall on WebQSP is lower than that on GrailQA. This is because  WebQSP has a smaller training set to learn from.

As for $k_{E1}$, when its value increases, the candidate entity recall generally drops. This is because an increase in $K_{E1}$ means to select more candidate entities based on popularity while fewer from those connected to the top retrieved relations but with lower popularity scores. 
Therefore, we default $k_{E1}$ at $1$ for GrailQA and $3$ for WebQSP, which yield the highest recall for the two datasets, respectively. 
Recall that we set the total number of candidate entities for each entity mention to 10 ($K_{E1} + K_{E2} = 10$), following our baselines (e.g., TIARA, RetinaQA, and Pangu). Therefore, we omit another study on $K_{E2}$, as it varies with $K_{E1}$.

\begin{table*}[h]
\small
\centering
\begin{tabular}{ll}
\toprule
\textbf{Question:} What is the name for the atomic units of length? \\ \midrule
\textbf{SpanMD (span classification):} \textcolor{red}{length} & (\ding{55})  \\ \midrule
\textbf{Ours} (extracting from generated logical form sketches):\\
\textbf{\hspace{6pt}Retrieved Relations:} measurement\_unit.measurement\_system.\underline{length\_units},\\
\hspace{87.5pt}measurement\_unit.time\_unit.measurement\_system, \\
\hspace{87.5pt}measurement\_unit.measurement\_system.time\_units... \\
\textbf{Generated Logical Form Sketch:}  (AND \textless{}class\textgreater~(JOIN \textless{}relation\textgreater~{[} \textcolor{blue}{atomic units} {]}))\hspace{10pt} &(\ding{51}) \vspace{1pt} \\ 
\hspace{6pt} \textbf{w/o SER:}  (AND \textless{}class\textgreater~(JOIN \textless{}relation\textgreater~{[} \textcolor{red}{length} {]}))\hspace{10pt} &(\ding{55})\\ 

\bottomrule
                                                                
\end{tabular}
\caption{Case study of entity mention detection by our model and SpanMD (a mention detection method commonly used by SOTA KBQA models) on the GrailQA validation set. The incorrect entity mention detected is colored in red, while the correct entity mention detected is colored in blue.}
\label{tab:md_case}
\end{table*}

\begin{figure*}[t]
    \centering
    \includegraphics{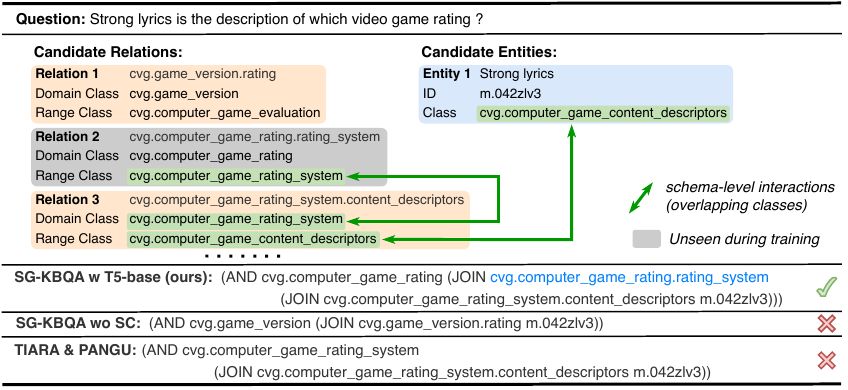}
    \caption{Case study of logical form generation by \model\ and two representative competitors TIARA and PANGU on the GrailQA validation set. Green arrows and boxes highlight schema-level interactions (overlapping classes) among KB elements.}
    \label{fig:lfg_case}
\end{figure*}

Figure~\ref{fig:kr_ke3} further shows the impact of $k_R$ and $k_{E3}$ -- recall that  $k_R$ is the number of top candidate relations considered, and $k_{E3}$ is the number of candidate entities matched for each entity mention. 
Now we show the F1 scores, as these parameters are used by 
our schema-guided logical form generation module. They directly affect the accuracy of the generated logical form and the corresponding question answers.

On GrailQA, increasing either $k_R$ or $k_{E3}$ leads to higher F1 scores, although the growth becomes marginal eventually. On WebQSP, the F1 scores peak at $k_R=25$ and $k_{E3}=4$. These results suggest that feeding an excessive number of candidate entities and relations to the logical form generator module has limited benefit. 
To avoid the extra computational costs (due to more input tokens) and to limit the input length for compatibility with smaller Seq2Seq models (e.g., T5-base), we use $k_R=20$ and $k_{E3}=2$ on both datasets.

\section{Model Running Time}\label{app:time}
\model\ takes 26 hours to train on the GrailQA dataset and 13.6 seconds to run inference for a test sample. It is faster on WebQSP which is a smaller dataset. Note that more than 10 hours of the training time were spent on the fallback logical form generation. If this step is skipped (which does not impact our model accuracy substantially as shown earlier), \model\ can be trained in about half a day. Another five hours were spent on fine-tuning the LLM for logical form generation, which can also be reduced by using a smaller model. 

As there is no full released code for the baseline models, it is infeasible to benchmark against them on model training time. For model  inference tests, TIARA has a partially released model (with a closed-source mention detection module). The model takes 11.4 seconds per sample (excluding the entity mention detection module) for inference on GrailQA, which is close to that of \model. Therefore, we have achieved a model that is more accurate than the baselines while being at least as fast in inference as one of the top performing baselines (i.e., TIARA+GAIN which shares the same inference procedure with TIARA).

\section{Case Study}\label{app:case_study}

To further show \model's generalizability to non-I.I.D. KBQA applications, we include a case study from the GrailQA validation set as shown in Table~\ref{tab:md_case} and Figure~\ref{fig:lfg_case}. 

\paragraph{Entity Mention Detection} 
Figure~\ref{tab:md_case} shows an entity mention detection example, comparing our entity detection module with SpanMD which is a mention detection method commonly used by SOTA KBQA models~\cite{shu_tiara_2022,ye_rng-kbqa_2022,faldu_retinaqa_2024}. In this case, SpanMD incorrectly detects \textsf{length} as an entity mention, which is actually part of the ground-truth relation (\textsf{measurement\_unit.$\ldots$.length\_units}) that is unseen in the training data. Our schema-guided entity retrieval module, on the other hand, leverages the retrieved relations as additional KB schema contexts to generate a logical form sketch.

By retrieving the unseen relation \textsf{measurement\_unit.measurement\_system.length\_units} from the question through our relation retrieval module, the model is provided with schema-level information not encountered during training. This enables it to accurately identify the correct entity mention, \textsf{atomic units}, from surrounding relation tokens---even when both the mention and relation are novel. Moreover, removing schema guidance (i.e., using a Seq2Seq model without SER) results in the same incorrect entity detection as SpanMD, highlighting the importance of schema-guidance in improving entity retrieval in non-I.I.D. scenarios.

\paragraph{Logical Form Generation}

Figure~\ref{fig:lfg_case} shows a logical form generation example.
Here, \model\ and TIARA (a representative generation-based model) and PANGU (a representative ranking-based SOTA model) have both retrieved the same sets of relations and entities in the retrieval stage which include false positives and unseen ground-truth relation. Meanwhile, the three models also share the same retrieved entity \textsf{m.042zlv3}. Despite accessing the correct entity and relevant relations, both TIARA and PANGU fail to generate the correct logical form, instead producing a logical form that consists of a relation composition seen at training. In contrast, \model successfully generates the correct logical form by leveraging schema context (i.e. the entity’s class as well as the domain and range classes of the retrieved relations which overlap). This enables \model to reason about the connectivity between the unseen ground-truth relation and the seen elements, demonstrating its superiority in generalizing to novel compositions of KB elements.

\section{Error Analysis}\label{app:error_analysis}
Following TIARA~\cite{shu_tiara_2022} and Pangu~\cite{gu_dont_2023}, we analyze 200 incorrect predictions randomly sampled from each of the GrailQA
validation set and the WebQSP test set where our model predictions are different from the ground truth. The errors of \model\ largely fall into the following three types:

\begin{table*}[t]
\centering
\small
\begin{tabular}{lccc}
\toprule
\textbf{Steps\textbackslash Models} & \textbf{SG-KBQA} & \textbf{TIARA} & \textbf{RnG-KBQA} \\
\midrule
Relation Retrieval & BERT-base-uncased & BERT-base-uncased & BERT-base-uncased \\
Logical Form Sketch Generation & T5-base & — & — \\
Entity Mention Detection & — & BERT-base-uncased & BERT-base-uncased \\
Entity Ranking & BERT-base-uncased & BERT-base-uncased & BERT-base-uncased \\
Logical Form Generation & LLaMA3.1-8B & T5-base & T5-base \\
\bottomrule
\end{tabular}
\caption{Backbone models used by \model and baselines at each pipeline step}
\label{tab:backbone_model}
\end{table*}

\begin{itemize}
    \item \textbf{Relation retrieval errors} (35\%). Failures in the relation retrieval step (e.g., failing to retrieve any ground-truth relations) can impinge the capability of our entity mention detection module to generate correct logical form sketches, which in turn leads to incorrect entity mention detection and entity retrieval. 

    \item \textbf{Entity retrieval errors} (32\%). Errors in the entity mentions generated by the logical form sketch parser can still occur even when the correct relations are retrieved, because some complex and unseen entity mentions require domain-specific knowledge. An example of such entity mentions is `\textsf{Non-SI units mentioned in the SI}', which refers to units that are not part of the International System (SI) of Units but are officially recognized for use alongside SI units. This entity mention involves two concepts that are very similar in their surface forms (\textsf{Non-SI} and \textsf{SI}). Without a thorough understanding of the  domain knowledge (\textsf{SI} standing for \textsf{International System of Units}), it is difficult for the entity mention detection module to identify the correct entity boundaries.

    \item \textbf{Logical form generation errors} (31\%). Generation of inaccurate or inexecutable logical forms can still occur when the correct entities and relations are retrieved. The main source of such errors is questions involving operators rarely seen in the training data (e.g., \textsf{ARGMIN} and \textsf{ARGMAX}). Additionally, there are highly ambiguous candidate entities that may confuse the model, leading to incorrect selections of entity-relation combinations. For example, for the question \textsf{Who writes twilight zone}, two candidate entities \textsf{m.04x4gj} and \textsf{m.0d\_rw} share the same entity name \textsf{twilight zone}. The former refers to a reboot of the TV series \textsf{The Twilight Zone} produced by Rod Serling and Michael Cassutt, while the latter is the original version of \textsf{The Twilight Zone} independently produced by Rod Serling. They share the same entity name and class (\textsf{tv.tv\_program}). There is insufficient contextual information for our logical form generator to  differentiate between the two. The generator eventually selected the higher-ranked entity which was incorrect, leading to producing an incorrect answer to the question \textsf{Rod Serling and Michael Cassutt}.

    \item The remaining errors (2\%) stem from incorrect annotations of comparative questions in the dataset. For example, \textsf{larger than} in a question is annotated as \textsf{LE} (less equal) in the ground-truth logical form.

\end{itemize}

\section{Retrieval Performance}\label{app:retrieval_performance}

\begin{table}[ht]
\small
\centering
\begin{tabular}{lcccc}
\toprule
\textbf{Model} & \textbf{Overall} & \textbf{I.I.D.} & \textbf{Comp.} & \textbf{Zero.} \\
\midrule
RnG-KBQA      & 80.4 & 86.6 & 83.3 & 76.5 \\
TIARA          & 85.4 & 91.3 & 86.9 & 82.2 \\
\textbf{SG-KBQA}        & \textbf{90.5} & \textbf{94.0} &\textbf{90.8} & \textbf{88.8} \\
\hspace{4pt}w/o SER    & 86.9 & 92.5 & 88.1 & 83.9 \\
\bottomrule
\end{tabular}
\caption{F1 scores of KBQA entity retrieval methods under different generalization scenarios.}
\label{tab:entity_retrieval}
\end{table}

\paragraph{Entity Retrieval} We report entity retrieval results under both I.I.D. and non-I.I.D. settings in Table~\ref{tab:entity_retrieval}, comparing our schema-guided entity retrieval module against baseline methods. Our method consistently outperforms the strongest baseline (TIARA), with larger performance gains observed in non-I.I.D. scenarios. Specifically, compared to a 2.7-point improvement in F1 score under the I.I.D. setting, our method achieves gains of 3.9 and 6.6 points under the compositional and zero-shot generalization settings, respectively. Furthermore, removing schema guidance from our logical form parser leads to F1 drops of 2.7 points in the compositional setting and 4.9 points in the zero-shot setting. These results further highlight the effectiveness of our schema-guided entity retrieval in enhancing overall model generalizability.

For ease of comparison, we summarize the backbone models used by \model and the baselines at each step of the pipeline in Table~\ref{tab:backbone_model}. In Section~\ref{sec:applicability}, we report results showing that even when using the same backbone model (T5-base) as the baselines (TIARA), our SG-KBQA still outperforms them.

\begin{figure}[h]
    \centering
    \includegraphics[width=\columnwidth]{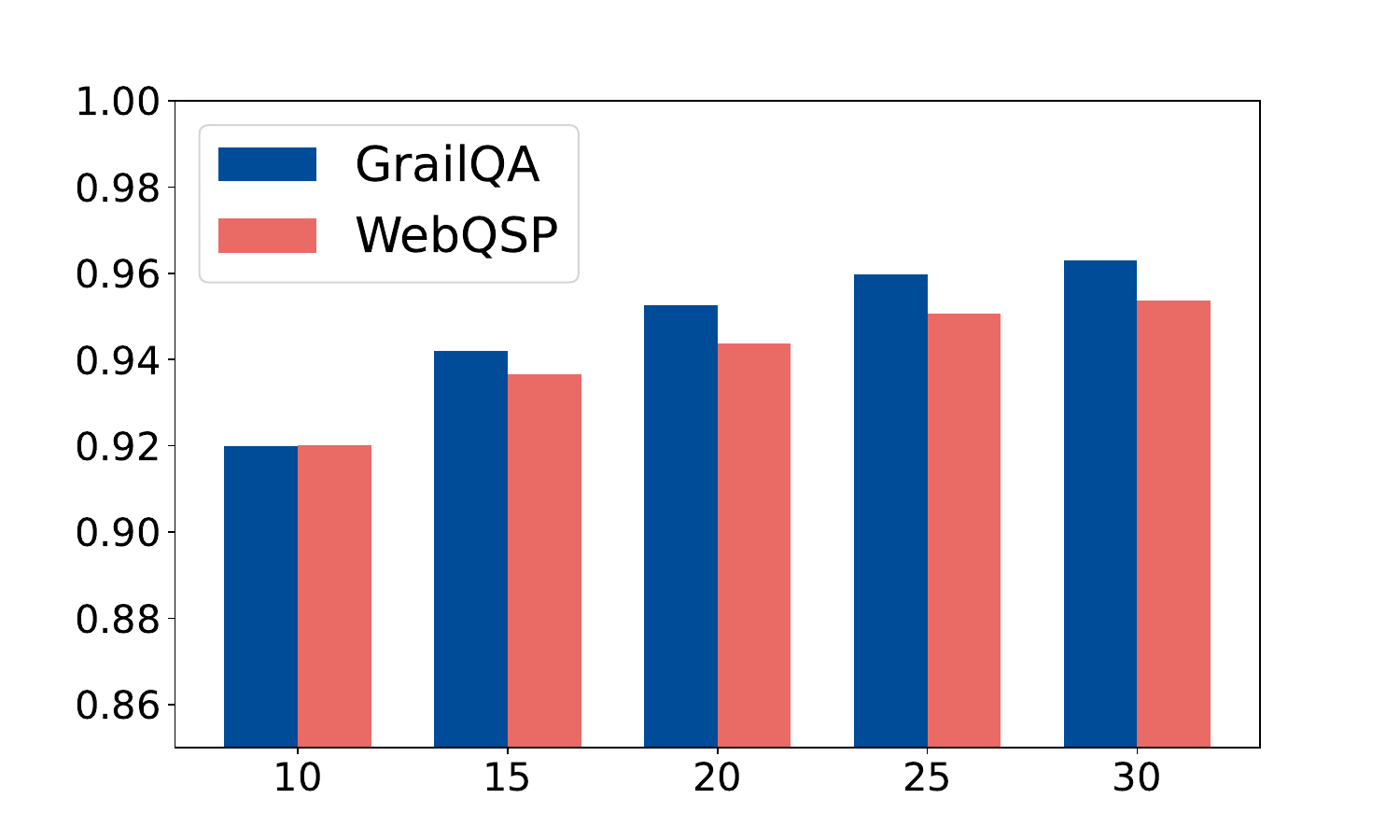}
    \caption{Recall of top-k relation retrieval on validation set of GrailQA and test set of WebQSP.}
    \label{fig:relation_retrieval}
\end{figure}

\paragraph{Relation Retrieval} Figure~\ref{fig:relation_retrieval} presents the recall of relation retrieval in SG-KBQA under different top-$k$ settings. Overall, as $k$ increases, the recall improves but with diminishing gains. Notably, even when $k=20$, the relation recall on the GrailQA validation set remains above 94\%, despite some question containing relations that were unseen during training. These results indicate that the cross-encoder-based retrieval model achieves high coverage of relevant relations, effectively retrieving both seen and unseen relations.

Although a larger $k$ introduces more noise (negative relations), our \model does not rely on highly precise relation retrieval. It leverages the top-20 retrieved relations to provide auxiliary context for the logical form sketch parser, and uses schema context to guide the LLM in selecting valid compositions of KB elements. Our experimental results show that \model is capable of identifying the correct relations and their compositions from a noisy candidate set, thereby improving robustness and generalization.

\end{document}